\documentclass[11pt, a4paper, logo, twocolumn]{berkeley}

\usepackage{moreverb,url}
\usepackage{array}    % for custom column types
\usepackage{tabularx} % for flexible-width table columns
\newcolumntype{Y}{>{\centering\arraybackslash}X}
\usepackage{afterpage}
\usepackage{amssymb}% http://ctan.org/pkg/amssymb
\usepackage{pifont}% http://ctan.org/pkg/pifont\
\usepackage{makecell}
\usepackage{array}
\usepackage{subcaption}
\newcolumntype{C}[1]{>{\centering\arraybackslash}b{#1}}

\newcommand\BibTeX{{\rmfamily B\kern-.05em \textsc{i\kern-.025em b}\kern-.08em
T\kern-.1667em\lower.7ex\hbox{E}\kern-.125emX}}

\usepackage[all]{hypcap}
\usepackage{xspace}
\usepackage[capitalize,noabbrev]{cleveref}
\usepackage{subcaption}
\usepackage{wrapfig}
\usepackage{lipsum}

\makeatletter
\def\mathcolor#1#{\@mathcolor{#1}}
\def\@mathcolor#1#2#3{%
  \protect\leavevmode
  \begingroup
    \color#1{#2}#3%
  \endgroup
}
\makeatother
\renewcommand{\today}{January 2024}

\usepackage{listings}
\newcommand{\bs}{\mathbf{s}}
\newcommand{\ba}{\mathbf{a}}
\newcommand{\states}{\mathcal{S}}
\newcommand{\actions}{\mathcal{A}}
\newcommand{\mdp}{\mathcal{M}}
\newcommand{\transition}{\mathcal{P}}
\newcommand{\reward}{r}

\newcommand{\ours}[1]{SERL\xspace}

\makeatletter
\let\blx@rerun@biber\relax
\makeatother

\title{SERL: A Software Suite for Sample-Efficient Robotic Reinforcement Learning}
\runningtitle{SERL: A Software Suite for Sample-Efficient Robotic Reinforcement Learning}

\author[1*]{Jianlan Luo}
\author[1*]{Zheyuan Hu}
\author[1]{Charles Xu}
\author[1]{You Liang Tan}
\author[2]{Jacob Berg}
\author[3]{Archit Sharma}
\author[4]{Stefan Schaal}
\author[3]{Chelsea Finn}
\author[2]{Abhishek Gupta}
\author[1]{Sergey Levine}

\affil[*]{Equal Contribution}
\affil[1]{Department of EECS, University of California, Berkeley}
\affil[2]{Department of Computer Science and Engineering, University of Washington}
\affil[3]{Department of Computer Science, Stanford University}
\affil[4]{Intrinsic Innovation LLC}

\correspondingauthor{Jianlan Luo and Zheyuan Hu, Berkeley AI Research Lab (BAIR), University of California, Berkeley: \href{mailto:jianlanluo@eecs.berkeley.edu}{jianlanluo@eecs.berkeley.edu}, \href{mailto:huzheyuan@berkeley.edu}{huzheyuan@berkeley.edu}. }

\begin{abstract}
In recent years, significant progress has been made in the field of robotic reinforcement learning (RL), enabling methods that handle complex image observations, train in the real world, and incorporate auxiliary data, such as demonstrations and prior experience.
However, despite these advances, robotic RL remains hard to use. It is acknowledged among practitioners that the particular implementation details of these algorithms are often just as important (if not more so) for performance as the choice of algorithm. We posit that a significant challenge to widespread adoption of robotic RL, as well as further development of robotic RL methods, is the comparative inaccessibility of such methods. To address this challenge, we developed a carefully implemented library containing a sample efficient off-policy deep RL method, together with methods for computing rewards and resetting the environment, a high-quality controller for a widely-adopted robot, and a number of challenging example tasks. We provide this library as a resource for the community, describe its design choices, and present experimental results. Perhaps surprisingly, we find that our implementation can achieve very efficient learning, acquiring policies for PCB board assembly, cable routing, and object relocation between 25 to 50 minutes
of training per policy on average, improving over state-of-the-art results reported for similar tasks in the literature. These policies achieve perfect or near-perfect success rates, extreme robustness even under perturbations, and exhibit emergent recovery and correction behaviors.
We hope that these promising results and our high-quality open-source implementation will provide a tool for the robotics community to facilitate further developments in robotic RL. Our code, documentation, and videos can be found at \mbox{\url{https://serl-robot.github.io/}}
\end{abstract}

\begin{document}
\teaserfigure{
    % \begin{figure*}[t]
    %     \centering
    %     \includegraphics[width=\linewidth]{images/Task.pdf}
    %     \caption{\footnotesize{Depiction of various tasks solved using \ours{} in the real world. These include PCB board insertion (left), cable routing (middle), and object relocation (right). \ours{} provides an out-of-the-box package for real-world reinforcement learning, with support for sample-efficient learning, learned rewards, and automation of resets.}}
    %     \label{fig:teaser}
    % \end{figure*}
    \centering
    \includegraphics[width=\textwidth]{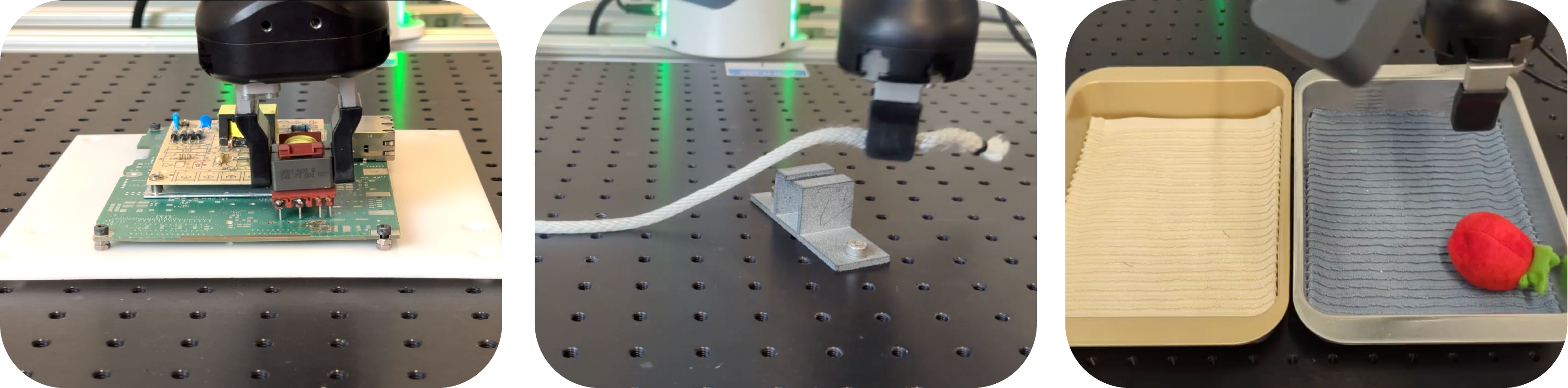}
    \captionof{figure}{\footnotesize{Depiction of various tasks solved using \ours{} in the real world. These include PCB board insertion (left), cable routing (middle), and object relocation (right). \ours{} provides an out-of-the-box package for real-world reinforcement learning, with support for sample-efficient learning, learned rewards, and automation of resets.}}
    \label{fig:teaser}
    \vspace{0.5cm}
}
\maketitle
\section{Introduction}\label{sec:intro}

Considerable progress on robotic reinforcement learning (RL) over the recent years has produced impressive results, with robots playing table tennis~\cite{buchler22tabletennis}, manipulating objects from raw images~\cite{gupta21mtrf, kalashnikov21mtopt, levine2016end}, grasping diverse objects~\cite{levine18grasping, mahler17dex}, and performing a wide range of other skills. However, despite the significant progress on the underlying algorithms, RL remains challenging to use for real-world robotic learning problems, and practical adoption has been more limited. We argue that part of the reason for this is that the implementation of RL algorithms, particularly for real-world robotic systems, presents a very large design space, and it is the challenge of navigating this design space, rather than limitations of algorithms \emph{per se}, that limit adoption. It is often acknowledged by practitioners in the field that details in the implementation of an RL algorithm might be as important (if not more important) as the particular choice of algorithm. Furthermore, real-world learning presents additional challenges with reward specification, implementation of environment resets, sample efficiency, compliant and safe control, and other difficulties that put even more stress on this issue. Thus, adoption and further research progress on real-world robotic RL may well be bottlenecked on \emph{implementation} rather than novel algorithmic innovations.

To address this challenge, our aim in this paper is to provide an open-source software framework, which we call \textbf{S}ample-\textbf{E}fficient \textbf{R}obotic reinforcement \textbf{L}earning (\ours{}), that aims to facilitate wider adoption of RL in real-world robotics. \ours{} consists of the following components: (1) a high-quality RL implementation that is geared towards real-world robotic learning and supports image observations and demonstrations; (2) implementations of several reward specification methods that are compatible with image observations, including classifiers and adversarial training; (3) support for learning ``forward-backward'' controllers that can automatically reset the task between trials~\cite{eysenbach18lnt}; (4) a software package that can in principle connect the aforementioned RL component to any robotic manipulator; and
%%SL.9.11: This sounds good, but let's temper this claim to be more realistic, so that ICRA reviewers don't reject the paper for overclaiming.
(5) an impedance controller design principle that is particularly effective for dealing with contact-rich manipulation tasks.
%%SL.9.11: Is the impedance controller only for Franka? If so we should make this clear (again to avoid giving the wrong impression)
%%JL 9.11 it's design choice that can be in principle made available to any torque controlled robots, I added "design principle" there 
%%SL.1.27: Are all of the statements in the above paragraph still true? (e.g., forward-backward, etc.)
%%ZH: the bin relocation experiment is reset-free and fwbw
Our aim in this paper is not to propose novel algorithms or methodology, but rather to offer a resource for the community to provide roboticists with a well-designed foundation both for future research on robotic RL, and other methods that might employ robotic RL as a subroutine. However, in the process of evaluating our framework, we also make a scientifically interesting empirical observation: when implemented properly in a carefully engineered software package, current sample-efficient robotic RL methods can attain very high success rates with relatively modest training times. The tasks in our evaluation are illustrated in Fig.~\ref{fig:teaser}: precise insertion tasks involving dynamic contact, deformable object manipulation with complex dynamics, and object relocation where the robot must learn without manually designed resets. %\todo{probably motivate more, cite some more papers saying RL is doing better on those tasks than other methods }
For each of these tasks, \ours{} is able to learn effectively within 15 - 60 min of training per policy (in terms of total wall-clock time), achieving near-perfect success rates, despite learning policies that operate on image observations. This result is significant because RL, particularly with deep networks and image inputs, is often considered to be highly inefficient. Our results challenge this assumption, suggesting careful implementations of existing techniques, combined with well-designed controllers and carefully selected components for reward specification and resets, can provide an overall system that is efficient enough for real-world use.
%We further demonstrate our framework on both the Franka Panda and the UR5 robotic arms, and compare our system to \todo{something}.

%\todo{Fill some gaps between}

%%SL.9.11: probably don't need this
%The rest of the paper is organized as follows, we explain the overall architecture and API of \ours{} in Sec.~\ref{sec:arch}; we then articulate the robot adapter as well as the impedance controller design; in Sec.~\ref{sec:exp} in addition to conducting experiments for aforementioned tasks; we also provide empirical ablations to verify the effectiveness of particular design choices.

%We hope with \ours{}. RL is going to become great again.

% \begin{figure*}[t]
% \captionsetup[subfigure]{labelformat=empty}
% \begin{subfigure}{1.0\textwidth}
% \centering
% \includegraphics[clip,width=0.5\textwidth]{example-image-a}
% \end{subfigure}%
% \newline
% \begin{subfigure}{1.0\textwidth}
% \centering
% \end{subfigure}%
% \vspace{-0.20cm}
% \caption{\textbf{Offline training tasks.} For each of the 11 training tasks, we run DDPGfD to collect data for offline training.}
% \label{fig:teaser}
% \end{figure*}
\label{introduction}
\section{Related Work}\label{sec:related_work}

While our framework combines existing RL methods into a complete robotic learning system, the particular combination of parts is carefully designed to provide for efficient and out-of-the-box reinforcement learning directly in the real world and, as shown in our experiments, achieves excellent results on a wide range of tasks. Here, we summarize both related prior methods and systems.

\noindent \textbf{Algorithms for real-world RL:}
Real-world robotic RL demands algorithms that are sample-efficient, can utilize onboard perception, and support easily specified rewards and resets. A number of algorithms have shown the ability to learn very efficiently directly in the real world~\cite{riedmiller2009reinforcement, westenbroek2022lyapunov, yang2020data, zhan2021framework, hou2020data, tebbe2021sample, popov2017data, luo2019reinforcement, zhao2022insertion, hu23reboot, residualrl, visual_residual_rl}, using variants of off-policy RL~\cite{kostrikov23wip, hu23reboot, luo2023rlif}, model-based RL~\cite{hester2013texplore, wu22daydreamer, nagabandi19pddm, rafailov21orllatent, luo2018deep}, and on-policy RL~\cite{zhu19dexterous}. These advances have been paired with advances in inferring rewards from raw visual observation through success classifiers ~\cite{fu2018variational, li21mural}, foundation-model-based rewards ~\cite{du2023vision, mahmoudieh22zeroshot, fan22minedojo}, and rewards from videos~\cite{ma23vip, ma23liv}. Additionally, to enable autonomous training, there have been a number of algorithmic advances in reset-free learning~\cite{gupta21mtrf, sharma2021Autonomous, zhu20r3l, xie22help, sharma23sir} that enable autonomous training with minimal human interventions. While these algorithmic advances are important, the contribution we make in this work is to provide a framework and software package to enable sample efficient reinforcement learning in the real world with a ready-made choice of methods that can work well for a variety of tasks. In doing so, we hope to lower the barrier of entry for new researchers to build better algorithms and training methodologies for robot learning in the real world.
%%SL.9.16: Most citations appear to be conflicted with us, need to cite non-conflicted RL work.
%% ZH: added more non-conflicted RL work

\noindent \textbf{Software packages for RL:}
There are a number of packages~\cite{d3rlpy, rlkit, stable-baselines, TFAgents} for RL, though to our knowledge, none aim to directly address real-world robotic RL specifically. \ours{} builds on the recently proposed RLPD algorithm, which is an off-policy RL algorithm with a high update-to-data ratio. \ours{} is not a library of RL algorithms for training agents in simulation, although it could be adapted to be so. Rather, \ours{} offers a full stack pipeline for robot control, going from low-level controllers to the interface for asynchronous and efficient training with an RL algorithm to additional machinery for inferring rewards and training without resets. In doing so, \ours{} provides an off-the-shelf package to help non-experts start using RL to train their physical robots in the real world, unlike prior libraries that aim to provide implementations of many methods -- that is, \ours{} offers a full ``vertical'' integration of components, whereas prior libraries focus on the ``horizontal.'' \ours{} is also not an RL benchmark package such as ~\cite{yu19meta, james20rlbench, mittal2023orbit}. \ours{} allows users to define their own tasks and success metrics directly in the real world, providing the software infrastructure for actually controlling and training robotic manipulators in these tasks. 

% d3rlpy
% jaxrl
% rlkit
% stable-baselines
% tfagents
% MBRL-Lib

\noindent \textbf{Software for real-world RL:}
There have been several previous packages that have proposed infrastructure for real world RL: for dexterous manipulation~\cite{ahn19robel}, tabletop furniture assembly~\cite{heo23furniture}, legged locomotion~\cite{kostrikov23wip}, and peg insertion~\cite{levine16gps}. These packages are effective in narrow situations, either using privileged information or training setups such as explicit tracking ~\cite{levine16gps, ahn19robel} or pure proprioception~\cite{kostrikov23wip}, or limited to imitation learning. In \ours{}, we show a full stack system that can be used for a wide variety of robotic manipulation tasks without requiring privileging of the training setups as in prior work. 

% Walk in the park
% R2D2?
% furniturebench
% montreal-franka
% shkruti group franka
% guided policy search
% PDDM
\label{related work}
\section{Preliminaries and Problem Statement}
\label{sec:background}

Robotic reinforcement learning tasks can be defined via an MDP $\mdp = \{\states, \actions, \rho, \transition, \reward, \gamma\}$, where $\bs \in \states$ is the state observation (e.g., an image in combination with the current end-effector position), $\ba \in \actions$ is the action (e.g., the desired end-effector pose), $\rho(\bs_0)$ is a distribution over initial states, $\transition$ is the unknown and potentially stochastic transition probabilities that depend on the system dynamics, and \mbox{$\reward: \states \times \actions \rightarrow \mathbb{R}$} is the reward function, which encodes the task. An optimal policy $\pi$ is one that maximizes the cumulative expected value of the reward, i.e., $E[\sum_{t=0}^\infty \gamma^t r(\bs_t,\ba_t)]$, where the expectation is taken with respect to the initial state distribution, transition probabilities, and policy $\pi$.

While the specification of the RL task is concise and simple, turning real-world robotic learning problems into RL problems requires care. First, the sample efficiency of the algorithm for learning $\pi$ is paramount: when the learning must take place in the real world, every minute and hour of training comes at a cost. Sample efficiency can be improved by using effective off-policy RL algorithms~\cite{konda99ac, haarnoja2018soft, fujimoto18td3}, but it can also be accelerated by incorporating prior data and demonstrations~\cite{Rajeswaran-RSS-18, ball2023efficient, nair2020accelerating}, which is important to achieve the fastest training times.

Additionally, many of the challenges with robotic RL lie beyond just the core algorithm for optimizing $\pi$. For example, the reward function $\reward$ might depend on image observations, and difficult for the user to specify manually.  Additionally, for episodic tasks where the robot resets to an initial state $\bs_0 \sim \rho(\bs_0)$ between trials, actually resetting the robot (and its environment) into one of these initial states is a mechanical operation that must somehow be automated.

Furthermore, the controller layer, which interfaces the MDP actions $\ba$ (e.g., end-effector poses) to the actual low-level robot controls, also requires great care, particularly for contact-rich tasks where the robot physically interacts with objects in the environment. Not only does this controller need to be accurate, but it must also be safe enough that the RL algorithm can explore with random actions during training. 

\ours{} will aim to provide ready-made solutions to each of these challenges, with a high-quality implementation of a sample-efficient off-policy RL method that can incorporate prior data, several choices for reward function specification, a forward-backward algorithm for learning resets, and a controller suitable for learning contact-rich tasks without damaging either the robot or objects in the environment.

%\begin{itemize}
%    \item Talk about RLPD with pretrained representation, why do we use RLPD, 
%    \item talk about reward function design, binary, ground truth, VICE
%    \item Talk about reset free (forward backward)
%\end{itemize}

\label{problem}
\section{Sample Efficient Robotic Reinforcement Learning in the Real-World}
\label{sec:arch}

Our software package, which we call \textbf{S}ample-\textbf{E}fficient \textbf{R}obotic reinforcement \textbf{L}earning (\ours{}), aims to make robotic RL in the real world accessible by providing ready-made solutions to the problems detailed in the previous section. This involves providing efficient vision-based reinforcement learning algorithms \emph{and} the infrastructure needed to support these learning algorithms for autonomous learning. We note that the purpose of such an endeavor is not to propose novel algorithms or tools, but rather to develop a software package that anyone can use easily for robotic learning, without complex setup procedures and painful integration across libraries. 
%This one-stop-shop for robotic learning is designed in a modular way to provide the software infrastructure to enable both users who want to use these algorithms as an off-the-shelf tool, as well as users who want to delve into the details and improve individual components of the system. 

% We provide an overview of the components of our system in Figure~\ref{fig:overview}.
%%SL.9.15: Create a figure that is kind of a block diagram of all the different parts of our system (see summary below).
The core reinforcement learning algorithm is derived from RLPD~\cite{ball2023efficient}, which itself is a variant of soft actor-critic~\cite{haarnoja2018soft}: an off-policy Q-function actor-critic method that can readily incorporate prior data (either suboptimal data or demonstrations) into the replay buffer for efficient learning. The reward functions can be specified either with a binary classifier or VICE~\cite{fu2018variational}, which provides a method to update the classifier during RL training with additional negatives from the policy. The reward function can also be specified by hand in cases where the robot state is sufficient to evaluate success (e.g., as in our PCB board assembly task). The resets can be provided via a forward-backward architecture~\cite{sharma2021Autonomous}, where the algorithm simultaneously trains two policies: a forward policy that performs the task, and a backward policy that resets the environment back to the initial state. On the robot system side, we also provide a universal adapter for interfacing our method to arbitrary robots, as well as an impedance controller that is particularly well-suited for contact-rich manipulation tasks.

\subsection{Core RL Algorithm: RLPD}
%%SL.9.15: My suggestion would be to actually present equations for the algorithm using notation in the preliminaries section

There are several desiderata for reinforcement learning algorithm to be deployed in this setting: (1) it must be efficient and able to make multiple gradient updates per time step, (2) it must be able to incorporate prior data easily and then continue improving with further experience, (3) it must be simple to debug and build on for new users. To this end, we build on the recently proposed RLPD~\cite{ball2023efficient} algorithm, which has shown compelling results on sample-efficient robotic learning. RLPD is an off-policy actor-critic reinforcement learning algorithm that builds on the success of temporal difference algorithms such as soft-actor critic~\cite{haarnoja2018soft}, but makes some key modifications to satisfy the desiderata above. RLPD makes three key changes: (i) high update-to-data ratio training (UTD), (ii) symmetric sampling between prior data and on-policy data, such that half of each batch comes from prior data and half from the online replay buffer, and (iii) layer-norm regularization during training. This method can train from scratch, or use prior data (e.g., demonstrations) to bootstrap learning. Each step of the algorithm updates the parameters of a parametric Q-function $Q_\phi(\bs,\ba)$ and actor $\pi_\theta(\ba|\bs)$ according to the gradient of their respective loss functions:
\begin{align*}
\mathcal{L}_Q(\phi) &\!=\! E_{\bs,\ba,\bs'}\!\!\left[ \!\left( Q_{\phi}(\bs,\ba) \!-\! \left( r(\bs,\ba) \!+\! \gamma E_{\ba'\sim \pi_{\theta}}[Q_{\bar{\phi}}(\bs',\ba')] \right) \right)^2\! \right]\\
\mathcal{L}_\pi(\theta) &\!=\! -E_{\bs}\left[ E_{\ba \sim \pi_\theta(\ba)}[Q_\phi(\bs,\ba)] + \alpha \mathcal{H}(\pi_\theta(\cdot | \bs) \right],
\end{align*}
\noindent where $Q_{\bar{\phi}}$ is a \emph{target network}~\cite{mnih2013playing}, and the actor loss uses entropy regularization with an adaptively adjusted weight $\alpha$~\cite{haarnoja2018soft}. Each update step uses a sample-based approximation of each expectation, with half of the samples drawn from the prior data (e.g., demonstrations), and half drawn from the \emph{replay buffer}~\cite{mnih2013playing}. For efficient learning, multiple update steps are performed per time step in the environment, which is referred to as the update-to-date (UTD) ratio, and regularizing the critic with layer normalization allows for higher UTD ratios and thus more efficient training~\cite{ball2023efficient}.
%Condition (1) ensures high data efficiency, since it increases the gradient throughput. This must be paired with condition (3) to prevent overfitting. Moreover, condition (2) ensures that the learned policies continue to improve on on-policy data collection, while not completely washing out the contributions from the prior expert data. While the algorithm of choice in our experiments is RLPD [cite], [alg name] supports other algorithms such as [TODO: list what it supports]. 

\subsection{Reward Specification with Classifiers}
\label{method:reward}
Reward functions are difficult to specify by hand when learning with image observations, as the robot typically requires some sort of perception system just to determine if the task was performed successfully. While some tasks, such as the PCB board assembly task in Fig.~\ref{fig:teaser}, can accommodate hand-specified rewards based on the location of the end effector (under the assumption that the object is held rigidly in the gripper),
%%SL.1.27: is this still how we actually do this task?
most tasks require rewards to be deduced from images. In this case, the reward function can be provided by a binary classifier that takes in the state observation $\bs$ and outputs the probability of a binary ``event'' $e$, corresponding to successful completion. The reward is then given by $r(\bs) = \log p(e|\bs)$.

%%SL.1.27: Do we have results for VICE in the experiments? If not, we should revise this section (or add it...)
This classifier can be trained either using hand-specified positive and negative examples, or via an adversarial method called VICE~\cite{fu2018variational}. The latter addresses a reward exploitation problem that can arise when learning with classifier based rewards, and removes the need for negative examples in the classifier training set: when the RL algorithm optimizes the reward $r(\bs) = \log p(e|\bs)$, it can potentially discover ``adversarial'' states that fool the classifier $p(e|\bs)$ to erroneously output high probabilities. VICE addresses this issue by adding all states visited by the policy into the training set for the classifier with negative labels, and updating the classifier after each iteration. In this way, the RL process is analogous to a generative adversarial network (GAN)~\cite{goodfellow2014generative}, with the policy acting as the generator and the reward classifier acting as the discriminator. Our framework thus supports all three types of rewards.

\subsection{Reset-Free Training with Forward-Backward Controllers}
When learning episodic tasks, the robot must reset the environment between task attempts. For example, when learning the object relocation task in Figure~\ref{fig:teaser}, each time the robot successfully moves the object to the target bin, it must then take it out and place it back into the initial bin. To remove the need for human effort in ``resets'', \ours{} supports ``reset-free'' training by using forward and backward controllers~\cite{han2015learning,gupta21mtrf}. In this setup, two policies are trained simultaneously using two independent RL agents, each with its own policy, Q-function, and reward function (specified via the methods in the previous section). The \emph{forward} agent learns to perform the task, and the \emph{backward} agent learns to return to the initial state(s). 
%In our system, both the forward and backward controllers are trained using the same RLPD algorithm, with the forward and backward policies taking turns to control the system, such that the final states of the forward policy are the initial states for the backward policy, and vice versa. 
While more complex reset-free training procedures can also be possible~\cite{gupta21mtrf}, we find that this simple recipe is sufficient for learning object manipulation tasks like the repositioning skill in Figure~\ref{fig:teaser}.

\subsection{Software Components}
\begin{figure}[t]
    \centering
    \includegraphics[width=\linewidth]{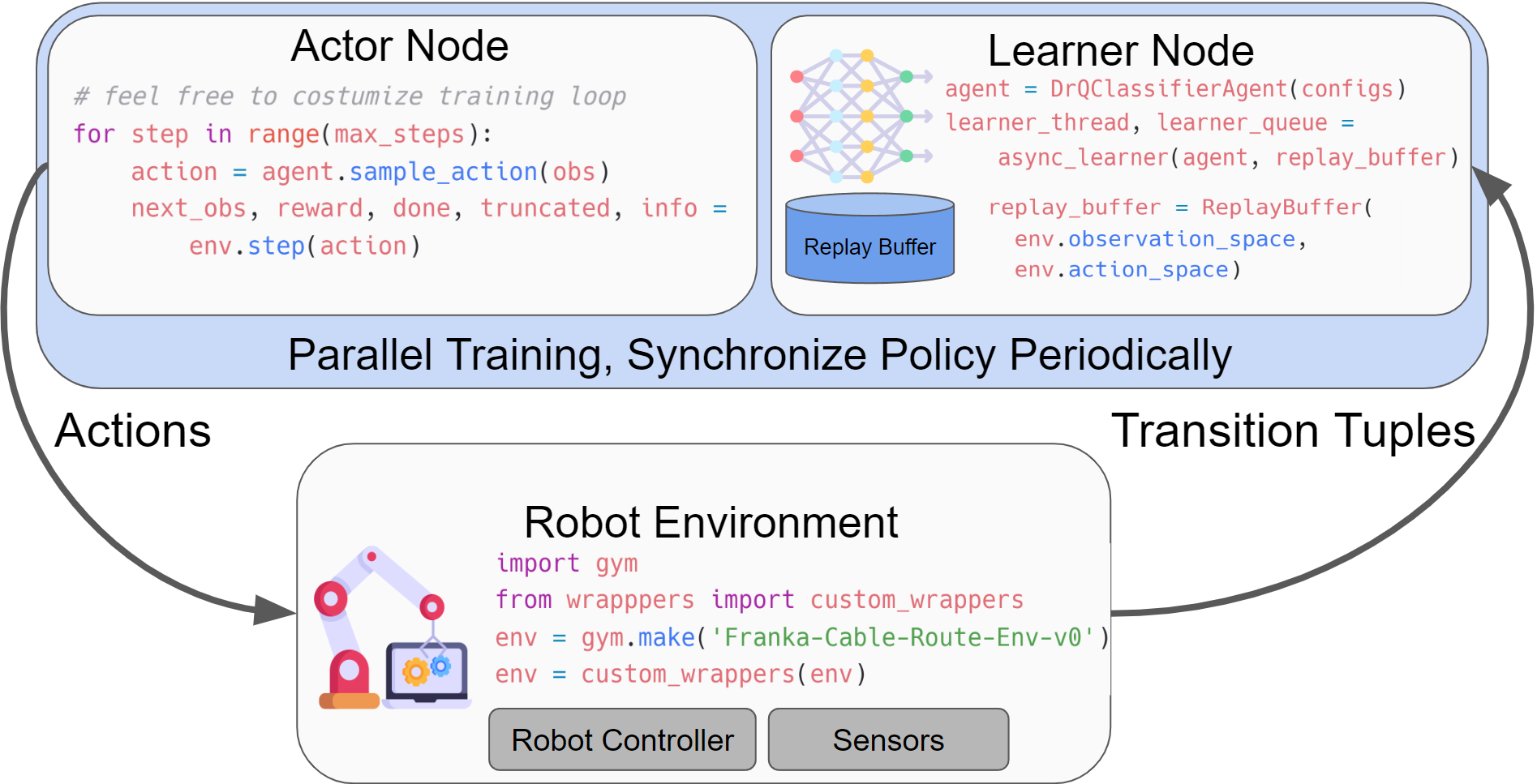}
    % \vspace{-0.5cm}
    \caption{\footnotesize{Software architecture and real-world robot training example code. \ours{} runs three parallel processes, consisting of the actor, which chooses actions, and the learner node, which actually runs the training code, and the robot environment, which executes the actions from the actor and contributes data back to the learner.}}
    \label{fig:code}
\end{figure}

\noindent \textbf{Environment adapters:} \ours{} aims to be easily usable for many robot environments. Although we provide a set of Gym environment wrappers and robot environments for the Franka arm as starter guides, users can also use their own existing environments or develop new environments as they see fit. Thus, the library does not impose additional constraints on the robot environment as long as it is Gym-like~\cite{brockman2016openai} as shown in Fig. \ref{fig:code}. We welcome contributions from the community to extend the support for readily deployable environment wrappers for other robots and tasks.

\noindent \textbf{Actor and learner nodes:} \ours{} includes options to train and act in parallel to decouple inferring actions and updating policies with a few lines of code as illustrated in Fig. \ref{fig:code}. We found this to be beneficial in sample-efficient real-world learning problems with high UTD ratios. By separating actor and learner on two different threads, \ours{} not only preserves the control frequency at a fixed rate, which is crucial for tasks that require immediate feedback and reactions, such as deformable objects and contact-rich manipulations, but also reduces the total wall-clock time spend training in the real world.

\subsection{Impedance Controller for Contact-Rich Tasks}
\label{controller}

%\section{CONTROLLER FOR CONTACT-RICH MANIPULATION TASKS}

Although our package should be compatible with any OEM robot controller, as described in Sec.~\ref{sec:arch}, we found that the choice of controllers can heavily affect the final performance. This is more pronounced for contact-rich manipulation. For example, in the PCB insertion task in Fig.~\ref{fig:teaser}, an overly stiff controller might bend the fragile pins and make insertion difficult, while an overly compliant controller might struggle to move the object into position quickly.

\begin{figure}
    \centering
    \includegraphics[width=\linewidth]{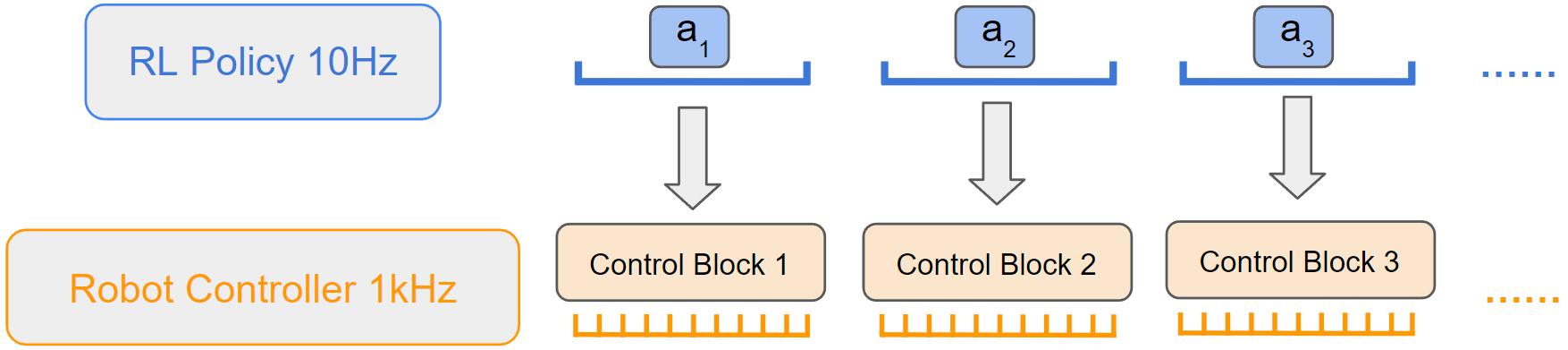}
    \caption{\footnotesize{A typical controller hierarchy for robotics RL. The output from the RL policy is tracked within a block of time by the downstream controller.}}
\vspace{-0.2cm}
    \label{fig:control_hierarchy}
\end{figure}

A typical setup for robotic RL employs a two-layered control hierarchy, where an RL policy produces set-point actions at a much lower frequency than the downstream real-time controller. The RL controller can set targets for the low-level controller to cause physically undesirable consequences. To illustrate this, let's consider the hierarchical controller structure presented in Fig.~\ref{fig:control_hierarchy}, where a high-level RL controller $\pi(\ba|\bs)$ sends control targets at 10HZ for the low-level impedance controller to track at 1K HZ, so one timestep from RL will block 100 timesteps of the low-level controller to execute.  A typical impedance control objective for this controller is 
\begin{equation*}
    F = k_p \cdot e + k_d \cdot \dot{e} + F_{ff} + F_{cor},
\end{equation*}
where $e = p - p_{ref}$, $p$ is the measured pose, and $p_{ref}$ is the target pose computed by the upstream controller, $F_{ff}$ is the feed-forward force, $F_{cor}$ is the Coriolis force, this objective will then be converted into joint space torques by multiplying Jacobian transpose and offset by nullspace torques. It acts like a spring-damper system around the equilibrium set by $p_{ref}$ with the stiffness coefficient being $k_p$ and the damping coefficient being $k_d$. As described above, this system will yield large forces if $p_{ref}$ is far away from the current pose, which can lead to a hard collision or damage when the arm is in contact with something. Therefore it's crucial to constrain the interaction force generated by it. However, directly reducing gains will hurt the controller's accuracy. Thus, we should bound $e$ so that $|e| \leq \Delta$,  and then the generated force from the spring-damper system will be bounded to $k_p \cdot |\Delta | + 2 k_d \cdot |\Delta| \cdot f$, $f$ is the control frequency. 

\begin{figure}[t]
    \centering
    \includegraphics[width=\linewidth]{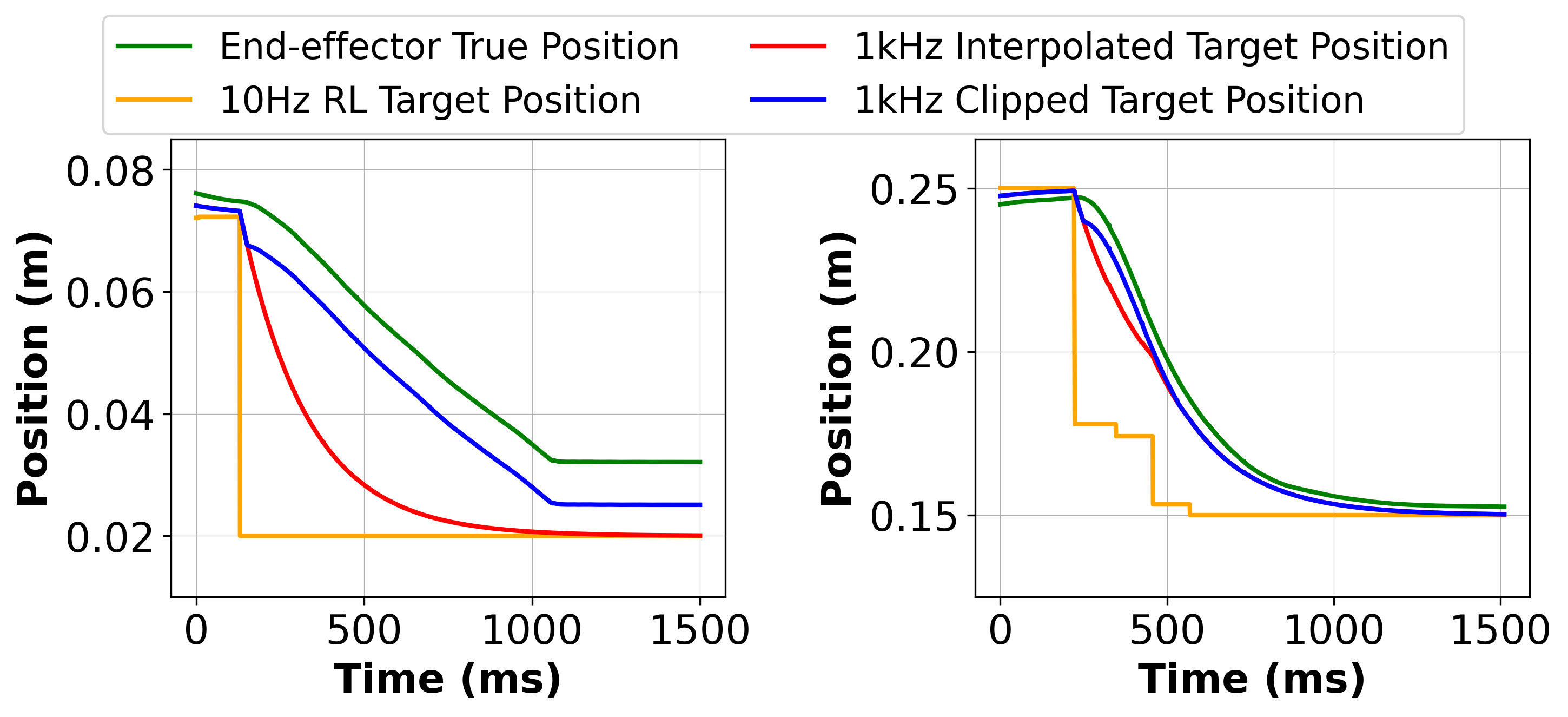}
    \caption{\footnotesize{Visualization of controller logs from the robot when commanded with different movements, for the z-axis of the end-effector. The orange line is the commanded target (the output of RL), red is the smoothed target sent to the real-time controller, blue is the clipped target, and green is the robot position after executing this controller. \textbf{Left:} The robot end-effector was commanded to move into contact with a hard surface and continue the movement despite the contact. The reference limiting mechanism clipped the target to avoid a hard collision. \textbf{Right:} The command is a fast free-space movement, which our reference limiting mechanism does \emph{not} block, allowing fast motion to the target.}}
    \label{fig:ref_limiting}
\end{figure}

One might wonder if we should directly clip the action output by the RL policy. This might seem reasonable, but can be impractical in some scenarios: some objects such as the PCB board may require a very small interaction force, implying a very small $\Delta$, usually on the order of micrometers; if the RL policy is only allowed to move at increments of micrometers, it would result in an extremely long learning process or very unstable training, because the episode would need enough time steps to allow the arm to move over long distances (e.g., approaching the insertion point). However, if we directly clip at the real-time layer, this will largely mitigate the issue without the need to constrain the RL policy to small actions. It will not block the free space movement of the RL policy as long as $M \cdot |\Delta| \geq |a|_{max}$, where $M$ is the number of control time-steps inside a block, as in Fig.~\ref{fig:control_hierarchy}. This value is usually large (e.g., $M=100$). 
%, it's usually a large number in this control hierarchy, e.g., M=100 in Fig.~\ref{}; $|a|_{max}$ is the maximum action the RL policy can output per time step;
At the same time, we strictly enforce the reference constraint at the real-time level whenever in contact. One might also wonder if it's possible to achieve the same result by using an external force/torque sensor. This may be undesirable for several reasons: (1) force/torque sensors can have significant noise, and obtaining the right hardware and calibration can be difficult; (2) even if we get such a threshold value, it's nontrivial to design robot motions to accommodate policy learning as well as obeying the force constraint. In practice, we found that clipping the reference in this way is simple but very effective, and is crucial to enable RL-based contact-rich manipulation tasks. We tested our controller on a Franka Panda robot and included the Franka Panda implementation with our package. However, this principle can be easily implemented on any torque-controlled robot. To verify the actual performance of the proposed controller, we report the actual tracking performance of moving the robot in free space and in contact with a table surface as in Fig.~\ref{fig:ref_limiting}, where we can see the controller indeed clamps the reference whenever in contact, while permitting fast movement in free space.\label{controller}

\begin{figure*}[ht!]
    \begin{center}
        \begin{minipage}{0.495\textwidth}
            \includegraphics[width=\linewidth]{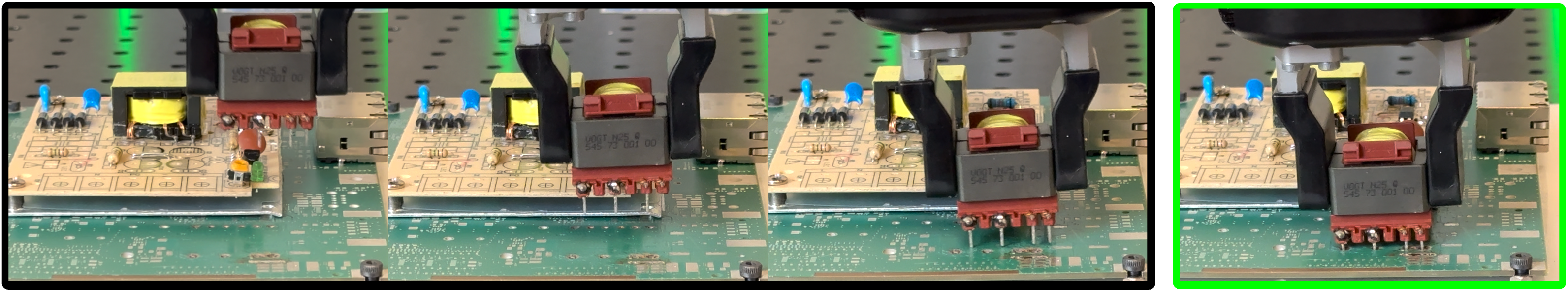}
        \end{minipage}
        \hfill
        \begin{minipage}{0.495\textwidth}
            \includegraphics[width=\linewidth]{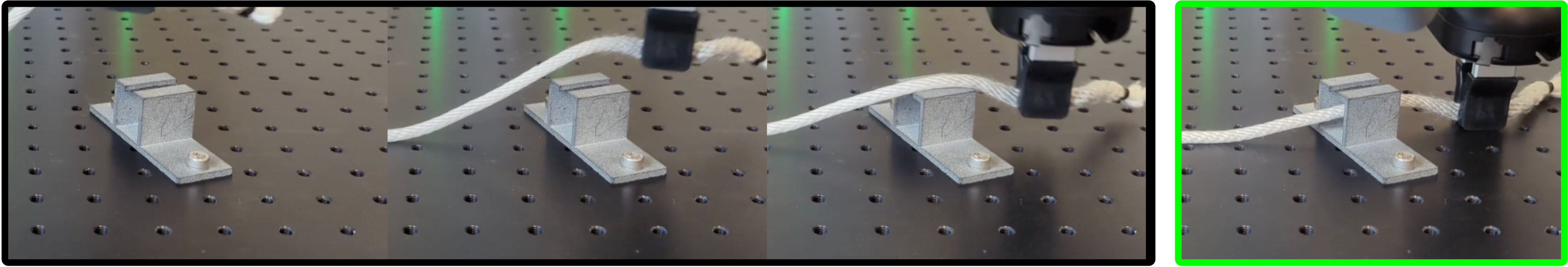}
        \end{minipage}

        \vspace{2.5mm} % Adjust this space as needed

        \begin{minipage}{0.495\textwidth}
            \includegraphics[width=\linewidth]{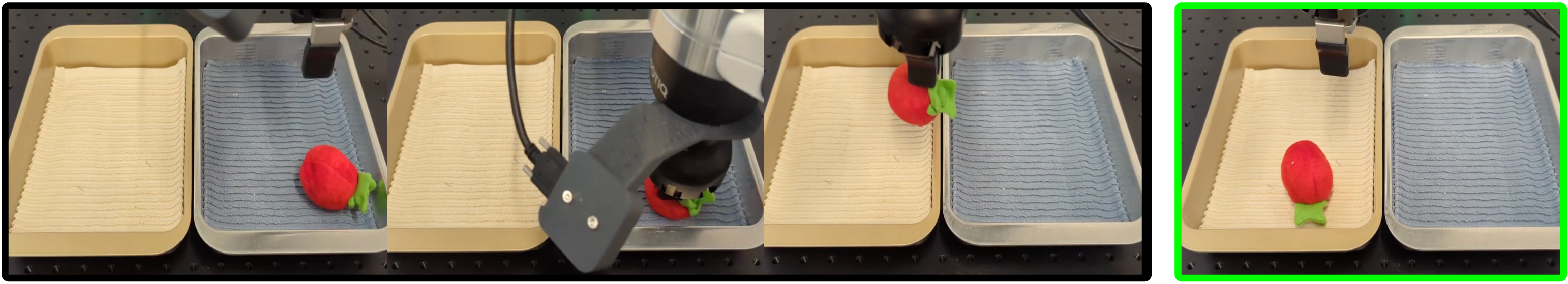}
        \end{minipage}
        \hfill
        \begin{minipage}{0.495\textwidth}
            \includegraphics[width=\linewidth]{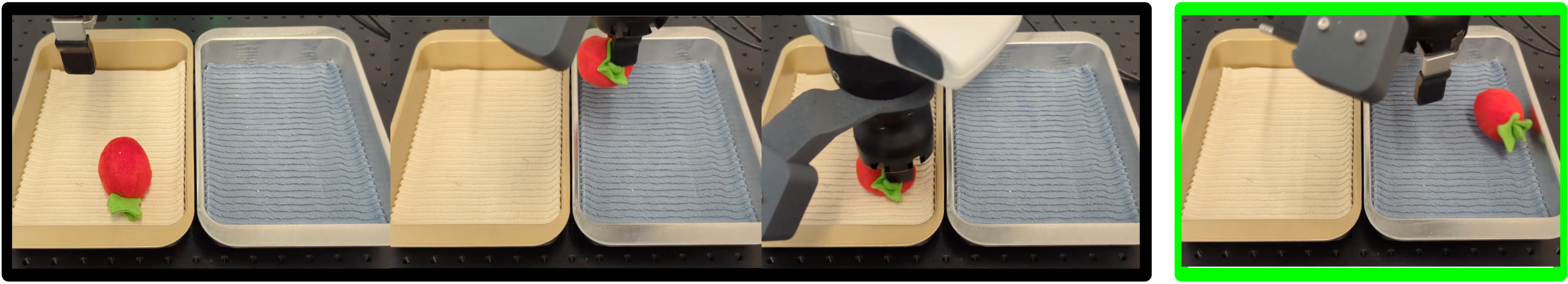}
        \end{minipage}

        \caption{
            \footnotesize{Illustration of the robot performing each task with our method: PCB Insertion (top left), Cable Routing (top right), Object Relocation - Forward (bottom left), and Object Relocation - Backward (bottom right). The green box indicates a state where the robot receives a high reward for completing the task. }
        }
        \label{fig:film_strips}
        \vspace{-0.5cm}
    \end{center}
\end{figure*}

\begin{table*}[t]
\centering
\resizebox{\textwidth}{!}{%
\begin{tabular}{c c c c c c c c}
\toprule
%%SL.1.27: can we use the name of the package rather than the author? it's pointless to say "Levine et al. (Levine et al.)" -- the parenthetical citation is always identical to the phrase right in front it. Instead, use the name of the package/method, so that people can figure out what you mean without having to look at the citation
%%ZH: fixing right now
\textbf{Package} & \textbf{Task} & \textbf{Training time} &\textbf{ Success rate }& \textbf{Demos} &  \textbf{Shaping?} & \textbf{Vision?} & \textbf{Open-sourced?}\\
\midrule
Guided Policy Search\cite{levine2016end} & Peg insertion & 3 hours & 70\% &  0 & Yes & Yes & Yes\\
DDPGfD  \cite{vecerik2018practical}&  Peg/clip insertion  &   1.5-2.5 hours     &     97\% / 77\% & 30 &No &  Yes      &   No     \\
Visual Residual RL \cite{visual_residual_rl} &  Connector insertion      &  Not mentioned     &    52\%  $\sim$  100\%     &    0 & Yes & Yes   &    No    \\
SHIELD \cite{Luo-RSS-21}&  Connector insertion  &   1.5 hours     &     99.8\%  & 25 & No &  Yes      &   No     \\
InsertionNet \cite{spector2021insertionnet}&  Connector insertion  &   40 mins     &     78.5\% - 100\% & 0 & Yes &  Yes      &   No     \\
\textbf{SERL (Ours)}  &  \textbf{PCB  Insertion}    &   \textbf{20 mins}    &    \textbf{100\%}   &  \textbf{20} & \textbf{No} & \textbf{Yes}     &   \textbf{Yes}     \\
\bottomrule
\end{tabular}
}
\caption{\footnotesize{Comparison to results reported on similar tasks in prior work. The overall success rates for our method are generally higher, and the training times are generally lower, as compared to prior results. Note also that the PCB board assembly task, shown in Figure~\ref{fig:teaser}, has very tight tolerances, likely significantly tighter than the coarser peg and connector insertion tasks studied in the prior works.}
\label{tab:comparisons}
}
\end{table*}

\subsection{Relative Observation and Action Frame}
\label{subsection:relativeframe}
The choice of action space is particularly important both for the ease of the RL training process and the ability of the learned policy to generalize to perturbations at test time. While \ours{} can operate on various action representations via a standard RL environment interface,
we found that a convenient mechanism of representing observations and actions in a relative coordinate system.

To develop an agent capable of adapting to a dynamic target, we propose a training procedure that simulates a moving target without the need for physical movement. 
The target, for instance, the PCB insertion socket holes, is fixed relative to the robot base frame, and the reward can be specified using any of the standard methods provided in Sec. \ref{method:reward}. 
At the beginning of each training episode, the pose of the robot's end-effector was randomized uniformly within a pre-defined area in the workspace. The robot's proprioceptive information is expressed with respect to frame of the end-effector's initial pose; the action output from the policy (6D twist) is relative to the current end-effector frame.
This is equivalent to physically moving the target when viewed relatively from the frame attached to the end-effector. 
More details on are described in the appendix \ref{subsection:appendix_relativeframe}. 
As a result, the policy can succeed even if the object moves or, as in some of our experiments, is perturbed in the middle of the episode. 

\label{methods}
\section{Experiments}\label{sec:exp}

\begin{table*}[t]
\centering
\resizebox{\textwidth}{!}{%
\begin{tabular}{c c c c c c c}
\toprule
\textbf{Task} & \textbf{\# of Demos} &  \textbf{Image Input} & \textbf{Random Reset} & \textbf{Reward Specification} & \textbf{Bin Size} & \textbf{Training Time}\\
\midrule
\textbf{PCB Component Insertion} & 20 & 2 wrist camera & True & Ground Truth & 10cm $\times$ 10cm & 20 mins \\
\textbf{Cable Routing} & 20 & 2 wrist camera & True & Binary Classifier & 20cm $\times$ 20cm & 31 mins \\
\textbf{Object Relocation (Forward-Backward)} & 20 & 1 wrist, 1 side camera & False & Binary Classifier & 20cm $\times$ 30cm & 105 mins \\
\bottomrule
\end{tabular}
}
%%SL.1.27: not sure why it's called a "bin size" -- initial end-effector position? can we think of some more logical name?
%%ZH: maybe workspace? or randomization box?
\caption{\footnotesize{Task parameters: During demo collection for both BC and RL, as well as online training, each episode's initial end-effector pose resets uniformly at random within a fixed region for the PCB and Cable task, while the free-floating object relocation task resets above the center of each bin.}}
\label{tab:comparisons}
\end{table*}

% \begin{table}[t]
% \centering
% \begin{tabular}{c c}
% \toprule
% \textbf{Task} &  \textbf{Training Time}\\
% \midrule
%  PCB Insertion & 20 min\\
% Object relocation& 120 min\\
% Cable Routing & 31 min \\
% \bottomrule
% \end{tabular}
% \caption{\footnotesize{Comparison to results reported on similar tasks in prior work. The overall success rates for our method are generally higher, and the training times are generally lower, as compared to prior results. Note also that the PCB board assembly task, shown in Figure~\ref{fig:teaser}, has very tight tolerances, likely significantly tighter than the coarser peg and connector insertion tasks studied in the prior works.}
% \label{tab:experiments}
% }
% \end{table}
Our experimental evaluation aims to study how efficiently our system can learn a variety of robotic manipulation tasks, including contact-rich tasks, deformable object manipulation, and free-floating object manipulation. These experiments demonstrate the breadth of applicability and efficiency of \ours{}. We use a Franka Panda arm and two wrist cameras attached to the end-effector to get close-in views.  Further details can be found at \mbox{\url{https://serl-robot.github.io/}}. We use an ImageNet pre-trained ResNet-10 ~\cite{he2015deep} as a vision backbone for the policy network and connect it to a 2-layer MLP. Observations include camera images and robot proprioceptive information such as end-effector pose, twist, force, and torque.
%%SL.1.27: I thought twist was the action, how is it also the observation?
%%ZH: this is the translation and angular velocity 
The policy outputs a 6D end-effector delta pose from the current pose, which is tracked by the low-level controller. The evaluation tasks are illustrated in Fig.~\ref{fig:film_strips} and described below:

%The experimental evaluation of our proposed system aims to answer the following questions:

%\begin{enumerate}
%    \item Can [system] learn manipulation tasks \emph{efficiently} from real-world experience?
%    \item Can [system] learn without requiring constant human intervention during training or environment instrumentation?
%    \item Can [system] learn generalizable behaviors with real-world experience? 
%\end{enumerate}

%\paragraph{Tasks:} Our experimental evaluation considers a number of different tasks involving both fixed objects and free objects. This diversity of tasks shows the ability of [system name] to learn in realistic environments, without requiring very contrived training setups. 

%%SL.9.15: IMPORTANT: somewhere we need to describe setup details, like pretrained conv stack, how many demos, how long it trains, etc.

\noindent \textbf{PCB insertion:} Inserting connectors into a PCB board demands fine-grained, contact-rich manipulation with sub-millimeter precision. This task is ideal for real-world training, as simulating and transferring such contact-rich interactions can be challenging. At the beginning of each training and evaluation episode, the initial end effector pose is sampled uniformly from a starting region, as described in Table~\ref{tab:comparisons}. %This task requires inserting connectors into a PCB board. It involves fine-grained and contact-rich manipulation, with sub-millimeter accuracy. This task is particularly well suited for real-world training because accurately modeling this type of contact-rich interaction and transferring this behavior from simulation can be very challenging.

\noindent \textbf{Cable routing:} This task involves routing a deformable cable into a clip's tight-fitting slot. This task requires the robot to perceive the cable and carefully manipulate it so that it fits into the clip while holding it at another location. This is particularly difficult for any method that relies on model-based control, or makes rigid-object assumptions, since both visual perception and handling of deformable objects is essential for access. Tasks of this sort often arise in manufacturing and maintenance scenarios. Similarly to the PCB task, the initial end effector pose is sampled uniformly within a starting region, as described in Table \ref{tab:comparisons}.%This task involves routing a pre-grasped deformable cable into the tight-fitting slot of a clip. The difficulty of this task comes from handling deformable objects with complex dynamics, closing the loop on visual perception. This task is representative of commonly occurring scenarios in manufacturing and maintenance, where a robot might need to route cables or hoses, and provides an interesting challenge for robotic manipulation.

\noindent \textbf{Object relocation:} This task requires moving a free-floating object between bins, requiring grasping and relocation. The intricacies of reward inference and reset-free training become especially pronounced in the manipulation of such free-floating objects. We define the forward task as picking up the object from the bin on the right side and placing it on the left, while the backward task moves the object back to the starting bin, undoing the forward task. %This task involves moving a free-floating object from one bin into another. This encompasses a broad range of pick-and-place manipulation tasks that may be encountered on deployment, where free-floating objects must be relocated from one position to another. This task is particularly interesting for real-world RL because the challenges of reward inference and reset-free training are exacerbated with free-floating object manipulation.

For each task, we initialize RL training from \textbf{20} teleoperated demonstrations using a Space Mouse. To confirm that demonstrations alone are insufficient to solve the task, we include a behavioral cloning (BC) baseline using \textbf{100} high-quality \textbf{expert} teleoperated demonstrations, roughly matching the total amount of data in the RL replay buffer when RL converges. Note that this is 5 times \emph{more} demonstrations than the amount provided by our method. Both RL and BC demonstrations are collected using the initial end effector randomization scheme described in Table~\ref{tab:comparisons}.
All training was done on a single Nvidia RTX 4090 GPU.

\begin{figure}[t]
    \centering
    \includegraphics[width=1\linewidth]{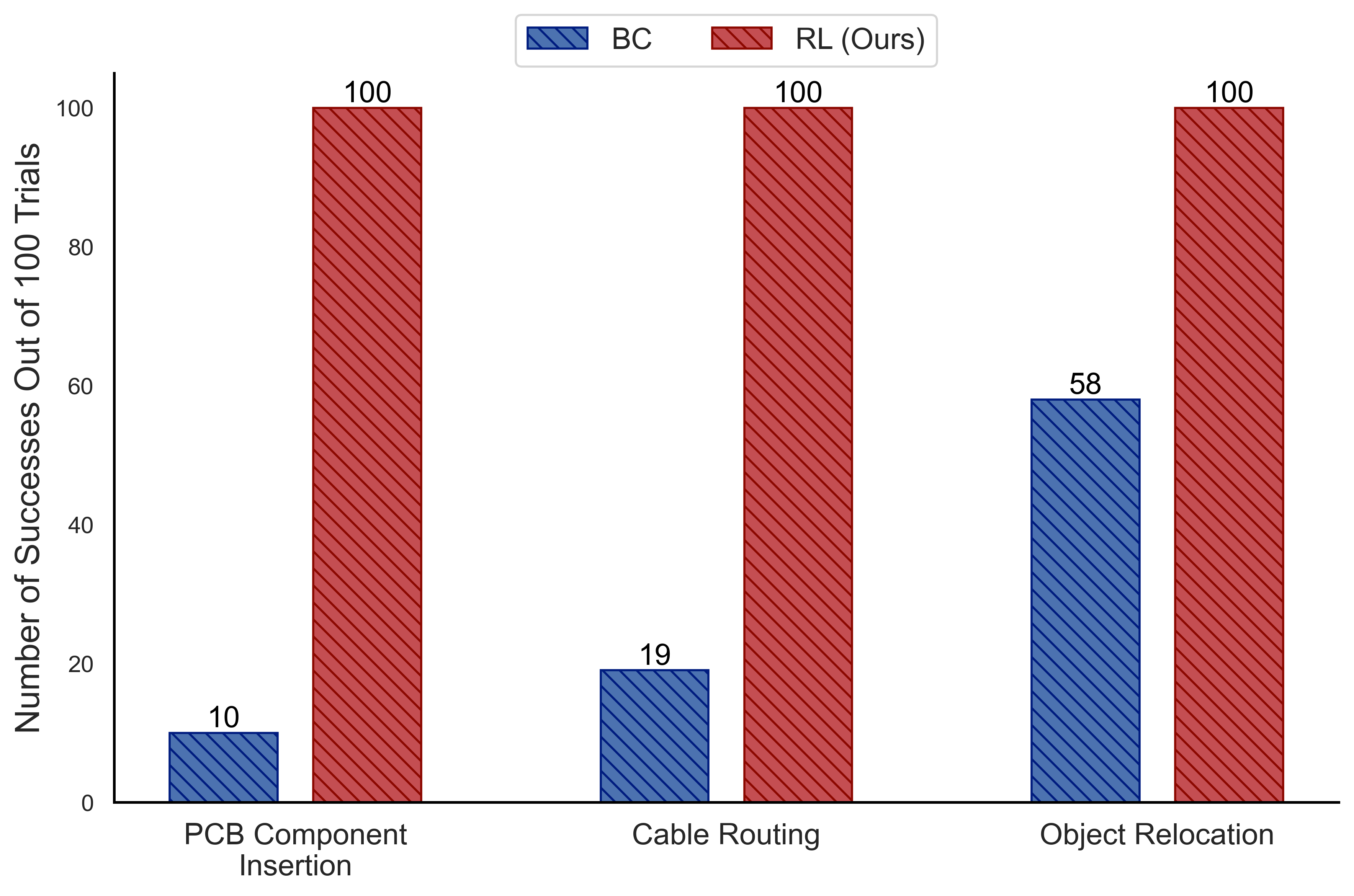}
    \caption{\footnotesize{Success rate comparisons: When evaluated for 100 trials per task, learned RL policies outperformed BC policies by a large margin, by \textbf{1.7x} for Object Relocation, by \textbf{5x} for Cable Routing, and by \textbf{10x} for PCB Insertion.}}
    \label{fig:success-rate}
\vspace{-0.2cm}
\end{figure}

\begin{figure}[t]
    \centering
    \includegraphics[width=1\linewidth]{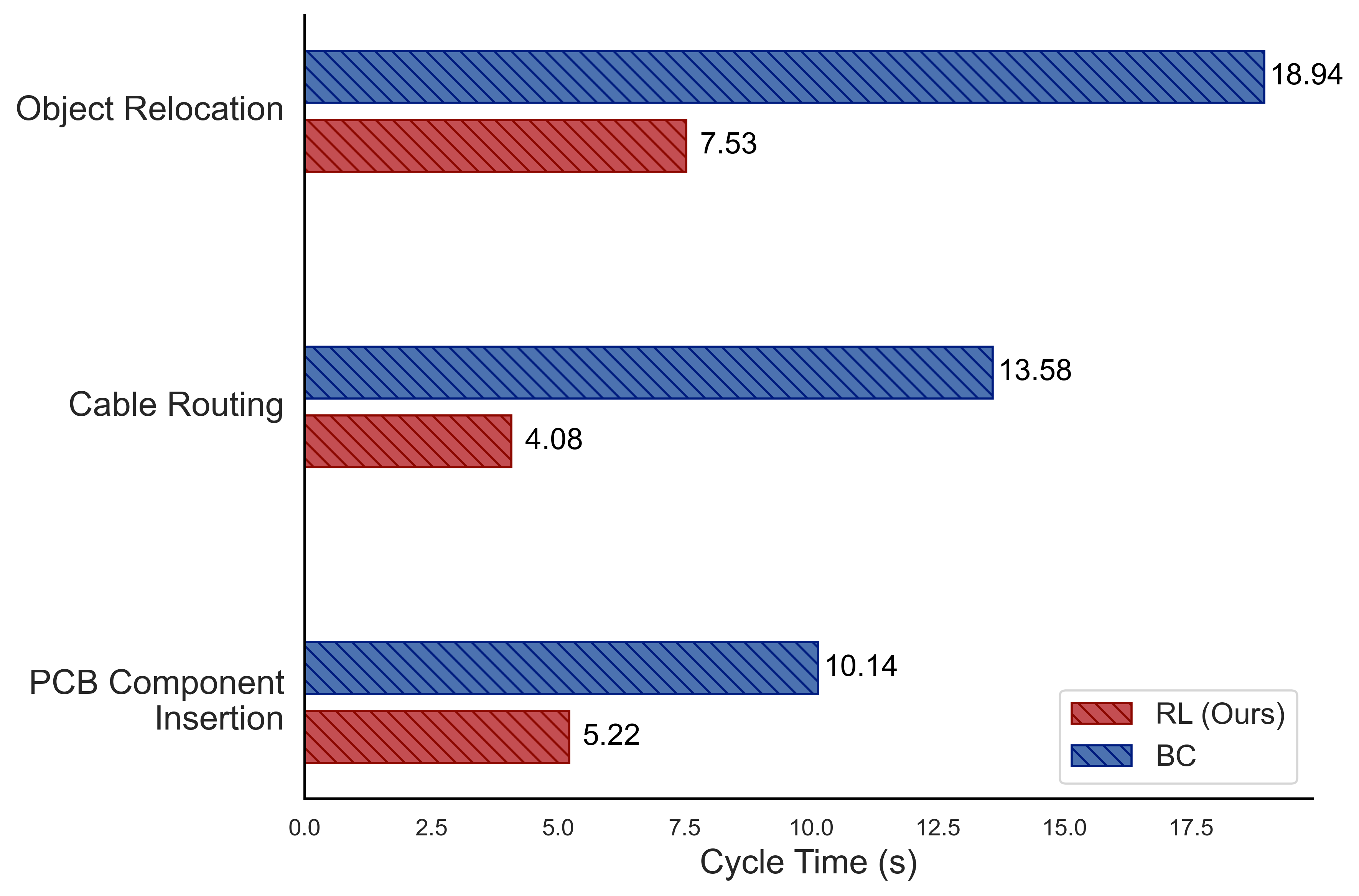}
    \caption{\footnotesize{Cycle time comparison: We recorded the average time taken for the robot to succeed in each task. RL policies are at least \textbf{2x} faster than BC policies trained with 100 high-quality human teleoperated demonstrations for all three tasks.}}
    \label{fig:cycle-time}
\vspace{-0.2cm}
\end{figure}

\paragraph{Results:} We report the results in Table~\ref{tab:comparisons}, and show example executions in Fig.~\ref{fig:film_strips}. We evaluated both BC and RL policies under the same conditions and protocols as detailed in Section~\ref{sec:exp}. Our RL policies achieve perfect success rates on all three tasks over all 100 trials. For the PCB insertion and cable routing task, our RL policies converge in under 30 minutes of real-world training, which includes all computation, resets, and intended stops. The free-floating object relocation task learns two policies (forward and backward), and total time amounts to less than an hour \emph{per policy}. For the cable routing task and PCB insertion task, our policies outperform BC baselines by a large margin, despite training with 5x fewer demonstrations than BC, suggesting that demos alone are insufficient. 
We report the results in terms of success rate and cycle time in Fig.~\ref{fig:success-rate} and Fig.~\ref{fig:cycle-time}. The learned RL policies not only outperformed their BC counterparts by as much as 10x in terms of success rate but also improved on the cycle time of the initial human demonstrations by up to 3x.
%%SL.1.27: I don't really know how to interpret Fig 8 or the above. I'm not sure it really adds much, maybe we can remove it to save on length?
%%ZH: moving to appendix for now

%This empirical finding resonates with prior convergence results in theory~\cite{watkins1998, jin2020}, we found it helpful for us to interpret the 100\% success rate achieved by RL empirically.
%%SL: commented out the above, it's kind of irrelevant (watkins and jin don't say anything about this setting, since we are not in a tabular or low-rank MDP)

\paragraph{Comparison to prior systems:} While it's difficult to directly compare our results to those of prior systems due to numerous differences in the setup, lack of consistently open-sourced code, and other discrepancies, we provide a summary of training times and success rates reported for tasks that are most similar to our PCB board insertion task in Table~\ref{tab:comparisons}. We chose this task because similar insertion or assembly tasks have been studied in prior work, and such tasks often present challenges with precision, compliant control, and sample efficiency. Compared to these prior works, our experiments do not use shaped rewards, which might require extensive engineering, though we do utilize a small amount of demonstration data (which some prior works eschew). The results reported in these prior works generally have either lower success rates or longer training times, or both, suggesting our implementation of sample-efficient RL matches or exceeds the performance of state-of-the-art methods in the literature, at least on this type of task. The closest performance to ours in the work of Spector et al.~\cite{spector2021insertionnet} includes a number of design decisions and inductive biases that are specific to insertion, whereas our method is generic and makes minimal task-specific assumptions. Although the components of our system are all based on (recent) prior work, the state-of-the-art performance of this combination illustrates our main thesis: the details of how deep RL methods are implemented can make a big difference.
%From the results above, we highlight one advantage of our package is the sample efficiency when dealing with visual inputs on these realistic dexterous manipulation tasks. Therefore, it's crucial to understand how is our software package different from others. Although it's hard to do a comparison exactly, we try our best to report on a fair basis. We pick insertion as a benchmark task since it's a widely studied problem in robotic manipulation, and we survey relevant open-sourced packages solving these tasks. We report the comparison in Table~\ref{}.

\begin{figure}[t]
\centering
\includegraphics[width=\columnwidth]{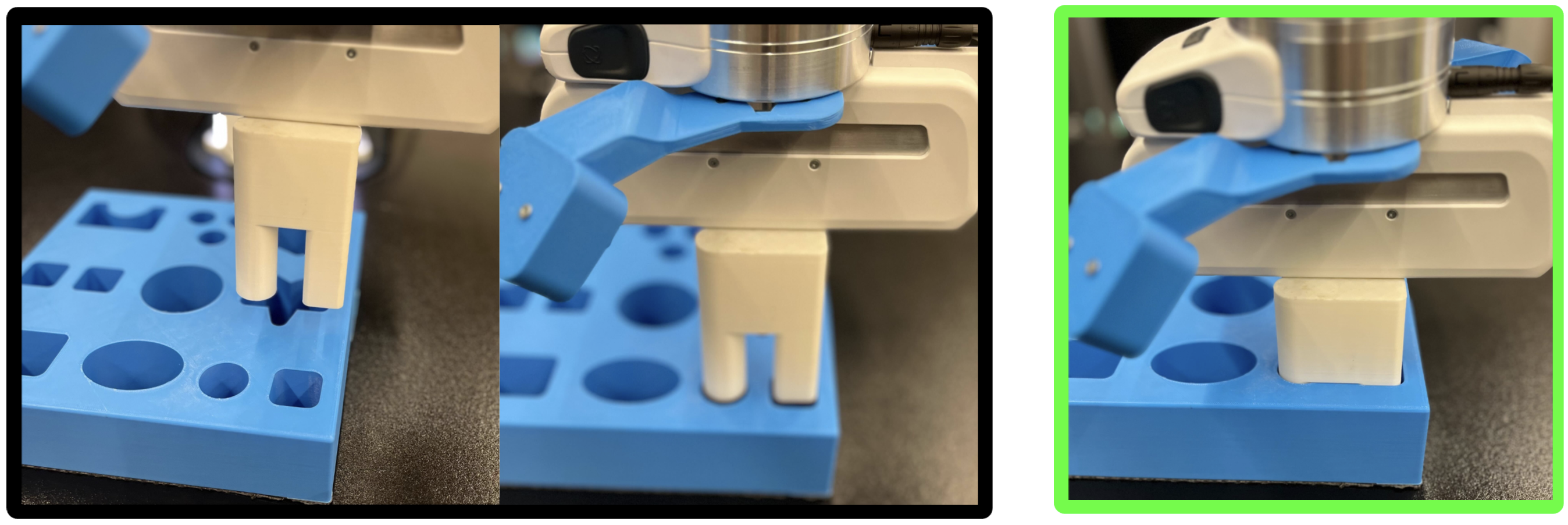}
\caption{Peg Insertion Task at University of Washington}
\label{fig:uwRobot}
\vspace{-0.2cm}
\end{figure}
\paragraph{Reproducibility:} A core mission of \ours{} is to lower the setup barriers and encourage reproducible robotic RL on different robot systems. To this end, we demonstrate the successful integration of \ours{} software suite on a robot arm operated at a different institution. 

Researchers at the University of Washington set up a Peg Insertion task using 3D printed parts from the Functional Manipulation Benchmark ~\cite{luo2024fmb} and used SERL to solve this challenging task. The overall preparation time including setting up the relevant hardware and software is less than 3 hours.
%The RL training used the same relative frame and randomization procedure as described in section \ref{subsection:relativeframe}. 
The policy converged in 19 minutes and achieved a 100/100 success rate with 20 initial human demonstrations, successfully reproducing our results.\label{experiments}
% \section{ANALYSIS}

% \subsection{How the amount of data affects pretraining performance}
% Ablate number of tasks, and compare adaptation and finetune performance
% [optional] Ablate number of data for each task, compare performance

% \subsection{Analyzing task latent space}
% Can latent space tell us about the functional difference between tasks? Does task close in latent space is also similar in our intuition? Does that relate to test task's accuracy?

% \subsection{Comparison with PEARL}
% Experiments to justify why we do not feed online interaction to the encoder

% % Is meta-RL only doing behavioral cloning? compare to BC

\label{experiments}
\section{Discussion}

We described a system and software package for robotic reinforcement learning that is aimed at making real-world RL more accessible both to researchers and practitioners. Our software package provides a carefully designed combination of ingredients for sample-efficient RL, automating reward design, automating environment resets with forward-backward controllers, and a controller framework that is particularly well-suited for contact-rich manipulation tasks. Furthermore, our experimental evaluation of our framework demonstrates that it can learn a range of diverse manipulation tasks very efficiently, with under an hour of training per policy when provided with a small number of demonstrations.
These results qualitatively compare well to state-of-the-art results in RL for manipulation in the literature, indicating that the particular choices in our framework are well-suited for obtaining very good real-world results even from image observations. 
Our framework does have a number of limitations. First, we do not aim to provide a comprehensive library with every possible RL method, 
%-- in fact, our goal is the opposite, to provide a framework that cuts down the number of necessary design choices and gives users a ready-made solution for robotic RL. 
%However, this does introduce some limitations: 
and some tasks and settings might be outside of our framework (e.g., non-manipulation tasks). Second, the full range of reward specifications and reset-free learning challenges still constitute an open problem in robotic RL research. Our classifier-based rewards and forward-backward controller might not be appropriate in every setting. Further research on these topics is needed to make robotic RL more broadly applicable. However, we hope that our software package will provide a reasonable ``default'' starting point for both researchers and practitioners wanting to experiment with real-world RL methods.

% , and this might in turn drive further development both in terms of applications of RL and more effective and efficient algorithms.
\label{conclusions}

\section*{Acknowledgments}This research was partially supported by Intrinsic Innovation LLC, the National Science Foundation under IIS-2150826, and ARO W911NF-21-1-0097. We would like to thank Rehaan Ahmad and Siri Gadipudi for participating in the discussion of this project.
% \clearpage
\section{Appendix}
\label{subsection:appendix_relativeframe}
\subsection{Details on Relative Observation and Action Frame}
%In many robotics applications, we want the RL policy to be able to solve the task across different poses, or even a moving target (i.e. route a cable through an unfixed clip or insert a connector into a moving socket). However, it is often hard to move the target during training, usually requiring a complex mechanism or a human supervisor to move the target object continuously. It also makes reward specification more challenging, making ground truth improbable and requiring more data to train learning-based reward signals. 

%To develop an agent capable of adapting to a dynamic target, we propose a training procedure that simulates a moving target without the need for physical movement. It achieves this purpose through randomizing the robot's end-effector in an instrumented manner instead, and using a relative coordinate frame system expressing actions and observations.

%The target is fixed relative to the robot base frame, and the reward can be specified using any of the standard methods provided in Sec. \ref{method:reward}. At the beginning of each training episode, the pose of the robot's end-effector was randomized uniformly across a pre-defined area in the workspace. 
%This is equivalent as physically moving the target when viewed relatively from the frame attached to the end-effector. 

%To ensure the policy does not simply memorize the fixed target pose, we transform the proprioceptive observation of the policy from the robot's base frame to the starting pose. 
%We now describe this coordinate system.
Let the robot's base frame be $\{s\}$; for the $i$-th episode of rolling out the policy, we denote $\{b_t^{(i)}\}$ as the end-effector frame expressed w.r.t. $\{s\}$ at a particular time step $t$; where $1\leq i \leq M$, $0\leq t \leq N$. For each episode, $\{b_0^{(i)}\}$ is sampled from a uniform distribution specifying the area of randomization. We want to express such proprioceptive information with respect to $\{b_0^{(i)}\}$. Thus, the policy will be applicable to a new location provided that the relative spatial distance between the robot's end-effector and the target remains consistent. This approach prevents overfitting to specific global locations within the reference frame $\{s\}$.
We achieve this by applying the following homogeneous transformation:
$$T_{b_0^{(i)}b_t^{(i)}} = T_{b_0^{(i)}}^{-1} \cdot T_{b_t^{(i)}}$$
where we use $T_{ab}$ to denote the homogeneous transformation matrix between frame $\{a\}$ and $\{b\}$. We feed the position and rotation information extracted from $T_{b_0^{(i)}b_t^{(i)}}$ to the policy.
Here we use $T_{ab}$ to denote the homogeneous transformation matrix between frame $\{a\}$ and $\{b\}$, defined as:

\begin{equation*}
  T_{ab} =   \begin{bmatrix}
R_{ab} & p_{ab} \\
0_{1 \times 3} & 1 
\end{bmatrix}.
\end{equation*}

The policy generates a six degrees of freedom (6 DoF) twist action, which is expressed in the reference frame from which it currently receives observations, i.e., $\{b_t^{(i)}\}$. 
Mathematically, the 6 DoF twist action $\mathcal{V}_t^{(i)}$ expressed in frame $\{b_t^{(i)}\}$ at timestep $t$. 
To interface with the robot's control software, which expects actions ${\mathcal{V}_t^{(i)}}^{\prime}$ expressed in the base frame $\{s\}$, we apply the Adjoint mapping:
$${\mathcal{V}_t^{(i)}}^{\prime} = {[\text{Ad}_t^{(i)}]}\mathcal{V}_t^{(i)}$$
where $[\text{Ad}_t^{(i)}]$ is a function of the homogeneous transformation $T_{b_0^{(i)}}$ defined as:
\begin{equation*}
  [\text{Ad}_t^{(i)}] =   \begin{bmatrix}
R_{b_t^{(i)}} & 0_{3 \times 3} \\
[p_{b_t^{(i)}}]\times R_{b_t^{(i)}} & R_{b_t^{(t)}}
\end{bmatrix}.
\end{equation*}

% \subsection{Visualizing the Q Function}
% \label{subsection:appendix_qheatmap}
% We also visualize the learned Q value heatmap for the PCB insertion task over time during one trial of policy learning in Fig.~\ref{fig:q_values}: we fix the z coordinate, and observe the changes in Q values over time for different state-action pairs in that surface. We find that the Q values converge as the learning proceeds, resulting in deterministic maximizing actions. 
% \begin{figure}[h]
%     \centering
%     \includegraphics[width=\linewidth]{images/heatmap_small.png}
%     \caption{\footnotesize{A visualization of learned Q values for the PCB Insertion task during training, we group states nearby into a grid, sample actions using the current policy, and report the average Q values.}}
%     \label{fig:q_values}
% \end{figure}
\label{appendix}
% \clearpage
\bibliography{ref}

\begin{thebibliography}{63}
\providecommand{\natexlab}[1]{#1}
\providecommand{\url}[1]{\texttt{#1}}
\expandafter\ifx\csname urlstyle\endcsname\relax
  \providecommand{\doi}[1]{doi: #1}\else
  \providecommand{\doi}{doi: \begingroup \urlstyle{rm}\Url}\fi

\bibitem[Ahn et~al.(2019)Ahn, Zhu, Hartikainen, Ponte, Gupta, Levine, and Kumar]{ahn19robel}
Michael Ahn, Henry Zhu, Kristian Hartikainen, Hugo Ponte, Abhishek Gupta, Sergey Levine, and Vikash Kumar.
\newblock {ROBEL:} robotics benchmarks for learning with low-cost robots.
\newblock In Leslie~Pack Kaelbling, Danica Kragic, and Komei Sugiura, editors, \emph{3rd Annual Conference on Robot Learning, CoRL 2019, Osaka, Japan, October 30 - November 1, 2019, Proceedings}, volume 100 of \emph{Proceedings of Machine Learning Research}, pages 1300--1313. {PMLR}, 2019.
\newblock URL \url{http://proceedings.mlr.press/v100/ahn20a.html}.

\bibitem[Ball et~al.(2023)Ball, Smith, Kostrikov, and Levine]{ball2023efficient}
Philip~J Ball, Laura Smith, Ilya Kostrikov, and Sergey Levine.
\newblock Efficient online reinforcement learning with offline data.
\newblock \emph{arXiv preprint arXiv:2302.02948}, 2023.

\bibitem[Brockman et~al.(2016)Brockman, Cheung, Pettersson, Schneider, Schulman, Tang, and Zaremba]{brockman2016openai}
Greg Brockman, Vicki Cheung, Ludwig Pettersson, Jonas Schneider, John Schulman, Jie Tang, and Wojciech Zaremba.
\newblock Openai gym.
\newblock \emph{arXiv preprint arXiv:1606.01540}, 2016.

\bibitem[B{\"{u}}chler et~al.(2022)B{\"{u}}chler, Guist, Calandra, Berenz, Sch{\"{o}}lkopf, and Peters]{buchler22tabletennis}
Dieter B{\"{u}}chler, Simon Guist, Roberto Calandra, Vincent Berenz, Bernhard Sch{\"{o}}lkopf, and Jan Peters.
\newblock Learning to play table tennis from scratch using muscular robots.
\newblock \emph{{IEEE} Trans. Robotics}, 38\penalty0 (6):\penalty0 3850--3860, 2022.
\newblock \doi{10.1109/TRO.2022.3176207}.
\newblock URL \url{https://doi.org/10.1109/TRO.2022.3176207}.

\bibitem[Du et~al.(2023)Du, Konyushkova, Denil, Raju, Landon, Hill, de~Freitas, and Cabi]{du2023vision}
Yuqing Du, Ksenia Konyushkova, Misha Denil, Akhil Raju, Jessica Landon, Felix Hill, Nando de~Freitas, and Serkan Cabi.
\newblock Vision-language models as success detectors.
\newblock \emph{arXiv preprint arXiv:2303.07280}, 2023.

\bibitem[Eysenbach et~al.(2018)Eysenbach, Gu, Ibarz, and Levine]{eysenbach18lnt}
Benjamin Eysenbach, Shixiang Gu, Julian Ibarz, and Sergey Levine.
\newblock Leave no trace: Learning to reset for safe and autonomous reinforcement learning.
\newblock In \emph{6th International Conference on Learning Representations, {ICLR} 2018, Vancouver, BC, Canada, April 30 - May 3, 2018, Conference Track Proceedings}. OpenReview.net, 2018.
\newblock URL \url{https://openreview.net/forum?id=S1vuO-bCW}.

\bibitem[Fan et~al.(2022)Fan, Wang, Jiang, Mandlekar, Yang, Zhu, Tang, Huang, Zhu, and Anandkumar]{fan22minedojo}
Linxi Fan, Guanzhi Wang, Yunfan Jiang, Ajay Mandlekar, Yuncong Yang, Haoyi Zhu, Andrew Tang, De{-}An Huang, Yuke Zhu, and Anima Anandkumar.
\newblock Minedojo: Building open-ended embodied agents with internet-scale knowledge.
\newblock In \emph{NeurIPS}, 2022.
\newblock URL \url{http://papers.nips.cc/paper\_files/paper/2022/hash/74a67268c5cc5910f64938cac4526a90-Abstract-Datasets\_and\_Benchmarks.html}.

\bibitem[Fu et~al.(2018)Fu, Singh, Ghosh, Yang, and Levine]{fu2018variational}
Justin Fu, Avi Singh, Dibya Ghosh, Larry Yang, and Sergey Levine.
\newblock Variational inverse control with events: A general framework for data-driven reward definition.
\newblock \emph{Advances in neural information processing systems}, 31, 2018.

\bibitem[Fujimoto et~al.(2018)Fujimoto, van Hoof, and Meger]{fujimoto18td3}
Scott Fujimoto, Herke van Hoof, and David Meger.
\newblock Addressing function approximation error in actor-critic methods.
\newblock In Jennifer~G. Dy and Andreas Krause, editors, \emph{Proceedings of the 35th International Conference on Machine Learning, {ICML} 2018, Stockholmsm{\"{a}}ssan, Stockholm, Sweden, July 10-15, 2018}, volume~80 of \emph{Proceedings of Machine Learning Research}, pages 1582--1591. {PMLR}, 2018.
\newblock URL \url{http://proceedings.mlr.press/v80/fujimoto18a.html}.

\bibitem[Goodfellow et~al.(2014)Goodfellow, Pouget-Abadie, Mirza, Xu, Warde-Farley, Ozair, Courville, and Bengio]{goodfellow2014generative}
Ian Goodfellow, Jean Pouget-Abadie, Mehdi Mirza, Bing Xu, David Warde-Farley, Sherjil Ozair, Aaron Courville, and Yoshua Bengio.
\newblock Generative adversarial nets.
\newblock \emph{Advances in neural information processing systems}, 27, 2014.

\bibitem[Guadarrama et~al.(2018)Guadarrama, Korattikara, Ramirez, Castro, Holly, Fishman, Wang, Gonina, Wu, Kokiopoulou, Sbaiz, Smith, Bartók, Berent, Harris, Vanhoucke, and Brevdo]{TFAgents}
Sergio Guadarrama, Anoop Korattikara, Oscar Ramirez, Pablo Castro, Ethan Holly, Sam Fishman, Ke~Wang, Ekaterina Gonina, Neal Wu, Efi Kokiopoulou, Luciano Sbaiz, Jamie Smith, Gábor Bartók, Jesse Berent, Chris Harris, Vincent Vanhoucke, and Eugene Brevdo.
\newblock {TF-Agents}: A library for reinforcement learning in tensorflow.
\newblock \url{https://github.com/tensorflow/agents}, 2018.
\newblock URL \url{https://github.com/tensorflow/agents}.
\newblock [Online; accessed 25-June-2019].

\bibitem[Gupta et~al.(2021)Gupta, Yu, Zhao, Kumar, Rovinsky, Xu, Devlin, and Levine]{gupta21mtrf}
Abhishek Gupta, Justin Yu, Tony~Z. Zhao, Vikash Kumar, Aaron Rovinsky, Kelvin Xu, Thomas Devlin, and Sergey Levine.
\newblock Reset-free reinforcement learning via multi-task learning: Learning dexterous manipulation behaviors without human intervention.
\newblock In \emph{{IEEE} International Conference on Robotics and Automation, {ICRA} 2021, Xi'an, China, May 30 - June 5, 2021}, pages 6664--6671. {IEEE}, 2021.
\newblock \doi{10.1109/ICRA48506.2021.9561384}.
\newblock URL \url{https://doi.org/10.1109/ICRA48506.2021.9561384}.

\bibitem[Haarnoja et~al.(2018)Haarnoja, Zhou, Abbeel, and Levine]{haarnoja2018soft}
Tuomas Haarnoja, Aurick Zhou, Pieter Abbeel, and Sergey Levine.
\newblock Soft actor-critic: Off-policy maximum entropy deep reinforcement learning with a stochastic actor.
\newblock In \emph{International conference on machine learning}, pages 1861--1870. PMLR, 2018.

\bibitem[Han et~al.(2015)Han, Levine, and Abbeel]{han2015learning}
Weiqiao Han, Sergey Levine, and Pieter Abbeel.
\newblock Learning compound multi-step controllers under unknown dynamics.
\newblock In \emph{2015 IEEE/RSJ International Conference on Intelligent Robots and Systems (IROS)}, pages 6435--6442. IEEE, 2015.

\bibitem[He et~al.(2015)He, Zhang, Ren, and Sun]{he2015deep}
Kaiming He, Xiangyu Zhang, Shaoqing Ren, and Jian Sun.
\newblock Deep residual learning for image recognition, 2015.

\bibitem[Heo et~al.(2023)Heo, Lee, Lee, and Lim]{heo23furniture}
Minho Heo, Youngwoon Lee, Doohyun Lee, and Joseph~J. Lim.
\newblock Furniturebench: Reproducible real-world benchmark for long-horizon complex manipulation.
\newblock In Kostas~E. Bekris, Kris Hauser, Sylvia~L. Herbert, and Jingjin Yu, editors, \emph{Robotics: Science and Systems XIX, Daegu, Republic of Korea, July 10-14, 2023}, 2023.
\newblock \doi{10.15607/RSS.2023.XIX.041}.
\newblock URL \url{https://doi.org/10.15607/RSS.2023.XIX.041}.

\bibitem[Hester and Stone(2013)]{hester2013texplore}
Todd Hester and Peter Stone.
\newblock Texplore: real-time sample-efficient reinforcement learning for robots.
\newblock \emph{Machine learning}, 90:\penalty0 385--429, 2013.

\bibitem[Hill et~al.(2018)Hill, Raffin, Ernestus, Gleave, Kanervisto, Traore, Dhariwal, Hesse, Klimov, Nichol, Plappert, Radford, Schulman, Sidor, and Wu]{stable-baselines}
Ashley Hill, Antonin Raffin, Maximilian Ernestus, Adam Gleave, Anssi Kanervisto, Rene Traore, Prafulla Dhariwal, Christopher Hesse, Oleg Klimov, Alex Nichol, Matthias Plappert, Alec Radford, John Schulman, Szymon Sidor, and Yuhuai Wu.
\newblock Stable baselines.
\newblock \url{https://github.com/hill-a/stable-baselines}, 2018.

\bibitem[Hou et~al.(2020)Hou, Fei, Deng, and Xu]{hou2020data}
Zhimin Hou, Jiajun Fei, Yuelin Deng, and Jing Xu.
\newblock Data-efficient hierarchical reinforcement learning for robotic assembly control applications.
\newblock \emph{IEEE Transactions on Industrial Electronics}, 68\penalty0 (11):\penalty0 11565--11575, 2020.

\bibitem[Hu et~al.(2024)Hu, Rovinsky, Luo, Kumar, Gupta, and Levine]{hu23reboot}
Zheyuan Hu, Aaron Rovinsky, Jianlan Luo, Vikash Kumar, Abhishek Gupta, and Sergey Levine.
\newblock {REBOOT:} reuse data for bootstrapping efficient real-world dexterous manipulation.
\newblock \emph{arXiv preprint arXiv:2309.03322}, 2024.

\bibitem[James et~al.(2020)James, Ma, Arrojo, and Davison]{james20rlbench}
Stephen James, Zicong Ma, David~Rovick Arrojo, and Andrew~J. Davison.
\newblock Rlbench: The robot learning benchmark {\&} learning environment.
\newblock \emph{{IEEE} Robotics Autom. Lett.}, 5\penalty0 (2):\penalty0 3019--3026, 2020.
\newblock \doi{10.1109/LRA.2020.2974707}.
\newblock URL \url{https://doi.org/10.1109/LRA.2020.2974707}.

\bibitem[Johannink et~al.(2018)Johannink, Bahl, Nair, Luo, Kumar, Loskyll, Ojea, Solowjow, and Levine]{residualrl}
Tobias Johannink, Shikhar Bahl, Ashvin Nair, Jianlan Luo, Avinash Kumar, Matthias Loskyll, Juan~Aparicio Ojea, Eugen Solowjow, and Sergey Levine.
\newblock Residual reinforcement learning for robot control.
\newblock \emph{CoRR}, abs/1812.03201, 2018.
\newblock URL \url{http://arxiv.org/abs/1812.03201}.

\bibitem[Kalashnikov et~al.(2021)Kalashnikov, Varley, Chebotar, Swanson, Jonschkowski, Finn, Levine, and Hausman]{kalashnikov21mtopt}
Dmitry Kalashnikov, Jacob Varley, Yevgen Chebotar, Benjamin Swanson, Rico Jonschkowski, Chelsea Finn, Sergey Levine, and Karol Hausman.
\newblock Mt-opt: Continuous multi-task robotic reinforcement learning at scale.
\newblock \emph{CoRR}, abs/2104.08212, 2021.
\newblock URL \url{https://arxiv.org/abs/2104.08212}.

\bibitem[Konda and Tsitsiklis(1999)]{konda99ac}
Vijay~R. Konda and John~N. Tsitsiklis.
\newblock Actor-critic algorithms.
\newblock In Sara~A. Solla, Todd~K. Leen, and Klaus{-}Robert M{\"{u}}ller, editors, \emph{Advances in Neural Information Processing Systems 12, {[NIPS} Conference, Denver, Colorado, USA, November 29 - December 4, 1999]}, pages 1008--1014. The {MIT} Press, 1999.
\newblock URL \url{http://papers.nips.cc/paper/1786-actor-critic-algorithms}.

\bibitem[Kostrikov et~al.(2023)Kostrikov, Smith, and Levine]{kostrikov23wip}
Ilya Kostrikov, Laura~M. Smith, and Sergey Levine.
\newblock Demonstrating {A} walk in the park: Learning to walk in 20 minutes with model-free reinforcement learning.
\newblock In Kostas~E. Bekris, Kris Hauser, Sylvia~L. Herbert, and Jingjin Yu, editors, \emph{Robotics: Science and Systems XIX, Daegu, Republic of Korea, July 10-14, 2023}, 2023.
\newblock \doi{10.15607/RSS.2023.XIX.056}.
\newblock URL \url{https://doi.org/10.15607/RSS.2023.XIX.056}.

\bibitem[Levine et~al.(2016{\natexlab{a}})Levine, Finn, Darrell, and Abbeel]{levine16gps}
Sergey Levine, Chelsea Finn, Trevor Darrell, and Pieter Abbeel.
\newblock End-to-end training of deep visuomotor policies.
\newblock \emph{J. Mach. Learn. Res.}, 17:\penalty0 39:1--39:40, 2016{\natexlab{a}}.
\newblock URL \url{http://jmlr.org/papers/v17/15-522.html}.

\bibitem[Levine et~al.(2016{\natexlab{b}})Levine, Finn, Darrell, and Abbeel]{levine2016end}
Sergey Levine, Chelsea Finn, Trevor Darrell, and Pieter Abbeel.
\newblock End-to-end training of deep visuomotor policies.
\newblock \emph{The Journal of Machine Learning Research}, 17\penalty0 (1):\penalty0 1334--1373, 2016{\natexlab{b}}.

\bibitem[Levine et~al.(2018)Levine, Pastor, Krizhevsky, Ibarz, and Quillen]{levine18grasping}
Sergey Levine, Peter Pastor, Alex Krizhevsky, Julian Ibarz, and Deirdre Quillen.
\newblock Learning hand-eye coordination for robotic grasping with deep learning and large-scale data collection.
\newblock \emph{Int. J. Robotics Res.}, 37\penalty0 (4-5):\penalty0 421--436, 2018.
\newblock \doi{10.1177/0278364917710318}.
\newblock URL \url{https://doi.org/10.1177/0278364917710318}.

\bibitem[Li et~al.(2021)Li, Gupta, Reddy, Pong, Zhou, Yu, and Levine]{li21mural}
Kevin Li, Abhishek Gupta, Ashwin Reddy, Vitchyr~H. Pong, Aurick Zhou, Justin Yu, and Sergey Levine.
\newblock {MURAL:} meta-learning uncertainty-aware rewards for outcome-driven reinforcement learning.
\newblock In Marina Meila and Tong Zhang, editors, \emph{Proceedings of the 38th International Conference on Machine Learning, {ICML} 2021, 18-24 July 2021, Virtual Event}, volume 139 of \emph{Proceedings of Machine Learning Research}, pages 6346--6356. {PMLR}, 2021.
\newblock URL \url{http://proceedings.mlr.press/v139/li21g.html}.

\bibitem[Luo et~al.(2018)Luo, Solowjow, Wen, Ojea, and Agogino]{luo2018deep}
Jianlan Luo, Eugen Solowjow, Chengtao Wen, Juan~Aparicio Ojea, and Alice~M Agogino.
\newblock Deep reinforcement learning for robotic assembly of mixed deformable and rigid objects.
\newblock In \emph{2018 IEEE/RSJ International Conference on Intelligent Robots and Systems (IROS)}, pages 2062--2069. IEEE, 2018.

\bibitem[Luo et~al.(2019)Luo, Solowjow, Wen, Ojea, Agogino, Tamar, and Abbeel]{luo2019reinforcement}
Jianlan Luo, Eugen Solowjow, Chengtao Wen, Juan~Aparicio Ojea, Alice~M Agogino, Aviv Tamar, and Pieter Abbeel.
\newblock Reinforcement learning on variable impedance controller for high-precision robotic assembly.
\newblock In \emph{2019 International Conference on Robotics and Automation (ICRA)}, pages 3080--3087. IEEE, 2019.

\bibitem[Luo et~al.(2021)Luo, Sushkov, Pevceviciute, Lian, Su, Vecerik, Ye, Schaal, and Scholz]{Luo-RSS-21}
Jianlan Luo, Oleg Sushkov, Rugile Pevceviciute, Wenzhao Lian, Chang Su, Mel Vecerik, Ning Ye, Stefan Schaal, and Jonathan Scholz.
\newblock {Robust Multi-Modal Policies for Industrial Assembly via Reinforcement Learning and Demonstrations: A Large-Scale Study}.
\newblock In \emph{Proceedings of Robotics: Science and Systems}, Virtual, July 2021.
\newblock \doi{10.15607/RSS.2021.XVII.088}.

\bibitem[Luo et~al.(2023)Luo, Dong, Zhai, Ma, and Levine]{luo2023rlif}
Jianlan Luo, Perry Dong, Yuexiang Zhai, Yi~Ma, and Sergey Levine.
\newblock Rlif: Interactive imitation learning as reinforcement learning.
\newblock \emph{arXiv preprint arXiv:2311.12996}, 2023.

\bibitem[Luo et~al.(2024)Luo, Xu, Liu, Tan, Lin, Wu, Abbeel, and Levine]{luo2024fmb}
Jianlan Luo, Charles Xu, Fangchen Liu, Liam Tan, Zipeng Lin, Jeffrey Wu, Pieter Abbeel, and Sergey Levine.
\newblock Fmb: a functional manipulation benchmark for generalizable robotic learning.
\newblock \emph{arXiv preprint arXiv:2401.08553}, 2024.

\bibitem[Ma et~al.(2023{\natexlab{a}})Ma, Kumar, Zhang, Bastani, and Jayaraman]{ma23liv}
Yecheng~Jason Ma, Vikash Kumar, Amy Zhang, Osbert Bastani, and Dinesh Jayaraman.
\newblock {LIV:} language-image representations and rewards for robotic control.
\newblock In Andreas Krause, Emma Brunskill, Kyunghyun Cho, Barbara Engelhardt, Sivan Sabato, and Jonathan Scarlett, editors, \emph{International Conference on Machine Learning, {ICML} 2023, 23-29 July 2023, Honolulu, Hawaii, {USA}}, volume 202 of \emph{Proceedings of Machine Learning Research}, pages 23301--23320. {PMLR}, 2023{\natexlab{a}}.
\newblock URL \url{https://proceedings.mlr.press/v202/ma23b.html}.

\bibitem[Ma et~al.(2023{\natexlab{b}})Ma, Sodhani, Jayaraman, Bastani, Kumar, and Zhang]{ma23vip}
Yecheng~Jason Ma, Shagun Sodhani, Dinesh Jayaraman, Osbert Bastani, Vikash Kumar, and Amy Zhang.
\newblock {VIP:} towards universal visual reward and representation via value-implicit pre-training.
\newblock In \emph{The Eleventh International Conference on Learning Representations, {ICLR} 2023, Kigali, Rwanda, May 1-5, 2023}. OpenReview.net, 2023{\natexlab{b}}.
\newblock URL \url{https://openreview.net/pdf?id=YJ7o2wetJ2}.

\bibitem[Mahler et~al.(2017)Mahler, Liang, Niyaz, Laskey, Doan, Liu, Ojea, and Goldberg]{mahler17dex}
Jeffrey Mahler, Jacky Liang, Sherdil Niyaz, Michael Laskey, Richard Doan, Xinyu Liu, Juan~Aparicio Ojea, and Ken Goldberg.
\newblock Dex-net 2.0: Deep learning to plan robust grasps with synthetic point clouds and analytic grasp metrics.
\newblock In Nancy~M. Amato, Siddhartha~S. Srinivasa, Nora Ayanian, and Scott Kuindersma, editors, \emph{Robotics: Science and Systems XIII, Massachusetts Institute of Technology, Cambridge, Massachusetts, USA, July 12-16, 2017}, 2017.
\newblock \doi{10.15607/RSS.2017.XIII.058}.
\newblock URL \url{http://www.roboticsproceedings.org/rss13/p58.html}.

\bibitem[Mahmoudieh et~al.(2022)Mahmoudieh, Pathak, and Darrell]{mahmoudieh22zeroshot}
Parsa Mahmoudieh, Deepak Pathak, and Trevor Darrell.
\newblock Zero-shot reward specification via grounded natural language.
\newblock In Kamalika Chaudhuri, Stefanie Jegelka, Le~Song, Csaba Szepesv{\'{a}}ri, Gang Niu, and Sivan Sabato, editors, \emph{International Conference on Machine Learning, {ICML} 2022, 17-23 July 2022, Baltimore, Maryland, {USA}}, volume 162 of \emph{Proceedings of Machine Learning Research}, pages 14743--14752. {PMLR}, 2022.
\newblock URL \url{https://proceedings.mlr.press/v162/mahmoudieh22a.html}.

\bibitem[Mittal et~al.(2023)Mittal, Yu, Yu, Liu, Rudin, Hoeller, Yuan, Tehrani, Singh, Guo, Mazhar, Mandlekar, Babich, State, Hutter, and Garg]{mittal2023orbit}
Mayank Mittal, Calvin Yu, Qinxi Yu, Jingzhou Liu, Nikita Rudin, David Hoeller, Jia~Lin Yuan, Pooria~Poorsarvi Tehrani, Ritvik Singh, Yunrong Guo, Hammad Mazhar, Ajay Mandlekar, Buck Babich, Gavriel State, Marco Hutter, and Animesh Garg.
\newblock Orbit: A unified simulation framework for interactive robot learning environments, 2023.

\bibitem[Mnih et~al.(2013)Mnih, Kavukcuoglu, Silver, Graves, Antonoglou, Wierstra, and Riedmiller]{mnih2013playing}
Volodymyr Mnih, Koray Kavukcuoglu, David Silver, Alex Graves, Ioannis Antonoglou, Daan Wierstra, and Martin Riedmiller.
\newblock Playing atari with deep reinforcement learning.
\newblock \emph{arXiv preprint arXiv:1312.5602}, 2013.

\bibitem[Nagabandi et~al.(2019)Nagabandi, Konolige, Levine, and Kumar]{nagabandi19pddm}
Anusha Nagabandi, Kurt Konolige, Sergey Levine, and Vikash Kumar.
\newblock Deep dynamics models for learning dexterous manipulation.
\newblock In Leslie~Pack Kaelbling, Danica Kragic, and Komei Sugiura, editors, \emph{3rd Annual Conference on Robot Learning, CoRL 2019, Osaka, Japan, October 30 - November 1, 2019, Proceedings}, volume 100 of \emph{Proceedings of Machine Learning Research}, pages 1101--1112. {PMLR}, 2019.
\newblock URL \url{http://proceedings.mlr.press/v100/nagabandi20a.html}.

\bibitem[Nair and Pong()]{rlkit}
Ashvin Nair and Vitchyr Pong.
\newblock rlkit.
\newblock \emph{Github}.
\newblock URL \url{https://github.com/rail-berkeley/rlkit}.

\bibitem[Nair et~al.(2020)Nair, Dalal, Gupta, and Levine]{nair2020accelerating}
Ashvin Nair, Murtaza Dalal, Abhishek Gupta, and Sergey Levine.
\newblock Accelerating online reinforcement learning with offline datasets, 2020.

\bibitem[Popov et~al.(2017)Popov, Heess, Lillicrap, Hafner, Barth-Maron, Vecerik, Lampe, Tassa, Erez, and Riedmiller]{popov2017data}
Ivaylo Popov, Nicolas Heess, Timothy Lillicrap, Roland Hafner, Gabriel Barth-Maron, Matej Vecerik, Thomas Lampe, Yuval Tassa, Tom Erez, and Martin Riedmiller.
\newblock Data-efficient deep reinforcement learning for dexterous manipulation.
\newblock \emph{arXiv preprint arXiv:1704.03073}, 2017.

\bibitem[Rafailov et~al.(2021)Rafailov, Yu, Rajeswaran, and Finn]{rafailov21orllatent}
Rafael Rafailov, Tianhe Yu, Aravind Rajeswaran, and Chelsea Finn.
\newblock Offline reinforcement learning from images with latent space models.
\newblock In Ali Jadbabaie, John Lygeros, George~J. Pappas, Pablo~A. Parrilo, Benjamin Recht, Claire~J. Tomlin, and Melanie~N. Zeilinger, editors, \emph{Proceedings of the 3rd Annual Conference on Learning for Dynamics and Control, {L4DC} 2021, 7-8 June 2021, Virtual Event, Switzerland}, volume 144 of \emph{Proceedings of Machine Learning Research}, pages 1154--1168. {PMLR}, 2021.
\newblock URL \url{http://proceedings.mlr.press/v144/rafailov21a.html}.

\bibitem[Rajeswaran et~al.(2018)Rajeswaran, Kumar, Gupta, Vezzani, Schulman, Todorov, and Levine]{Rajeswaran-RSS-18}
Aravind Rajeswaran, Vikash Kumar, Abhishek Gupta, Giulia Vezzani, John Schulman, Emanuel Todorov, and Sergey Levine.
\newblock Learning complex dexterous manipulation with deep reinforcement learning and demonstrations.
\newblock In \emph{Proceedings of Robotics: Science and Systems}, Pittsburgh, Pennsylvania, June 2018.
\newblock \doi{10.15607/RSS.2018.XIV.049}.

\bibitem[Riedmiller et~al.(2009)Riedmiller, Gabel, Hafner, and Lange]{riedmiller2009reinforcement}
Martin Riedmiller, Thomas Gabel, Roland Hafner, and Sascha Lange.
\newblock Reinforcement learning for robot soccer.
\newblock \emph{Autonomous Robots}, 27:\penalty0 55--73, 2009.

\bibitem[Schoettler et~al.(2020)Schoettler, Nair, Luo, Bahl, Aparicio~Ojea, Solowjow, and Levine]{visual_residual_rl}
Gerrit Schoettler, Ashvin Nair, Jianlan Luo, Shikhar Bahl, Juan Aparicio~Ojea, Eugen Solowjow, and Sergey Levine.
\newblock Deep reinforcement learning for industrial insertion tasks with visual inputs and natural rewards.
\newblock In \emph{2020 IEEE/RSJ International Conference on Intelligent Robots and Systems (IROS)}, pages 5548--5555, 2020.
\newblock \doi{10.1109/IROS45743.2020.9341714}.

\bibitem[Seno and Imai(2022)]{d3rlpy}
Takuma Seno and Michita Imai.
\newblock d3rlpy: An offline deep reinforcement learning library.
\newblock \emph{Journal of Machine Learning Research}, 23\penalty0 (315):\penalty0 1--20, 2022.
\newblock URL \url{http://jmlr.org/papers/v23/22-0017.html}.

\bibitem[Sharma et~al.(2021)Sharma, Xu, Sardana, Gupta, Hausman, Levine, and Finn]{sharma2021Autonomous}
Archit Sharma, Kelvin Xu, Nikhil Sardana, Abhishek Gupta, Karol Hausman, Sergey Levine, and Chelsea Finn.
\newblock Autonomous reinforcement learning: Benchmarking and formalism.
\newblock \emph{arXiv preprint arXiv:2112.09605}, 2021.

\bibitem[Sharma et~al.(2023)Sharma, Ahmed, Ahmad, and Finn]{sharma23sir}
Archit Sharma, Ahmed~M. Ahmed, Rehaan Ahmad, and Chelsea Finn.
\newblock Self-improving robots: End-to-end autonomous visuomotor reinforcement learning.
\newblock \emph{CoRR}, abs/2303.01488, 2023.
\newblock \doi{10.48550/arXiv.2303.01488}.
\newblock URL \url{https://doi.org/10.48550/arXiv.2303.01488}.

\bibitem[Spector and Castro(2021)]{spector2021insertionnet}
Oren Spector and Dotan~Di Castro.
\newblock Insertionnet -- a scalable solution for insertion, 2021.

\bibitem[Tebbe et~al.(2021)Tebbe, Krauch, Gao, and Zell]{tebbe2021sample}
Jonas Tebbe, Lukas Krauch, Yapeng Gao, and Andreas Zell.
\newblock Sample-efficient reinforcement learning in robotic table tennis.
\newblock In \emph{2021 IEEE international conference on robotics and automation (ICRA)}, pages 4171--4178. IEEE, 2021.

\bibitem[Vecerik et~al.(2018)Vecerik, Sushkov, Barker, Rothörl, Hester, and Scholz]{vecerik2018practical}
Mel Vecerik, Oleg Sushkov, David Barker, Thomas Rothörl, Todd Hester, and Jon Scholz.
\newblock A practical approach to insertion with variable socket position using deep reinforcement learning, 2018.

\bibitem[Westenbroek et~al.(2022)Westenbroek, Castaneda, Agrawal, Sastry, and Sreenath]{westenbroek2022lyapunov}
Tyler Westenbroek, Fernando Castaneda, Ayush Agrawal, Shankar Sastry, and Koushil Sreenath.
\newblock Lyapunov design for robust and efficient robotic reinforcement learning.
\newblock \emph{arXiv preprint arXiv:2208.06721}, 2022.

\bibitem[Wu et~al.(2022)Wu, Escontrela, Hafner, Abbeel, and Goldberg]{wu22daydreamer}
Philipp Wu, Alejandro Escontrela, Danijar Hafner, Pieter Abbeel, and Ken Goldberg.
\newblock Daydreamer: World models for physical robot learning.
\newblock In Karen Liu, Dana Kulic, and Jeffrey Ichnowski, editors, \emph{Conference on Robot Learning, CoRL 2022, 14-18 December 2022, Auckland, New Zealand}, volume 205 of \emph{Proceedings of Machine Learning Research}, pages 2226--2240. {PMLR}, 2022.
\newblock URL \url{https://proceedings.mlr.press/v205/wu23c.html}.

\bibitem[Xie et~al.(2022)Xie, Tajwar, Sharma, and Finn]{xie22help}
Annie Xie, Fahim Tajwar, Archit Sharma, and Chelsea Finn.
\newblock When to ask for help: Proactive interventions in autonomous reinforcement learning.
\newblock In \emph{NeurIPS}, 2022.
\newblock URL \url{http://papers.nips.cc/paper\_files/paper/2022/hash/6bf82cc56a5fa0287c438baa8be65a70-Abstract-Conference.html}.

\bibitem[Yang et~al.(2020)Yang, Caluwaerts, Iscen, Zhang, Tan, and Sindhwani]{yang2020data}
Yuxiang Yang, Ken Caluwaerts, Atil Iscen, Tingnan Zhang, Jie Tan, and Vikas Sindhwani.
\newblock Data efficient reinforcement learning for legged robots.
\newblock In \emph{Conference on Robot Learning}, pages 1--10. PMLR, 2020.

\bibitem[Yu et~al.(2019)Yu, Quillen, He, Julian, Hausman, Finn, and Levine]{yu19meta}
Tianhe Yu, Deirdre Quillen, Zhanpeng He, Ryan Julian, Karol Hausman, Chelsea Finn, and Sergey Levine.
\newblock Meta-world: {A} benchmark and evaluation for multi-task and meta reinforcement learning.
\newblock In Leslie~Pack Kaelbling, Danica Kragic, and Komei Sugiura, editors, \emph{3rd Annual Conference on Robot Learning, CoRL 2019, Osaka, Japan, October 30 - November 1, 2019, Proceedings}, volume 100 of \emph{Proceedings of Machine Learning Research}, pages 1094--1100. {PMLR}, 2019.
\newblock URL \url{http://proceedings.mlr.press/v100/yu20a.html}.

\bibitem[Zhan et~al.(2021)Zhan, Zhao, Pinto, Abbeel, and Laskin]{zhan2021framework}
Albert Zhan, Ruihan Zhao, Lerrel Pinto, Pieter Abbeel, and Michael Laskin.
\newblock A framework for efficient robotic manipulation.
\newblock In \emph{Deep RL Workshop NeurIPS 2021}, 2021.

\bibitem[Zhao et~al.(2022)Zhao, Luo, Sushkov, Pevceviciute, Heess, Scholz, Schaal, and Levine]{zhao2022insertion}
Tony~Z. Zhao, Jianlan Luo, Oleg Sushkov, Rugile Pevceviciute, Nicolas Heess, Jon Scholz, Stefan Schaal, and Sergey Levine.
\newblock Offline meta-reinforcement learning for industrial insertion.
\newblock In \emph{2022 International Conference on Robotics and Automation (ICRA)}, pages 6386--6393, 2022.
\newblock \doi{10.1109/ICRA46639.2022.9812312}.

\bibitem[Zhu et~al.(2019)Zhu, Gupta, Rajeswaran, Levine, and Kumar]{zhu19dexterous}
Henry Zhu, Abhishek Gupta, Aravind Rajeswaran, Sergey Levine, and Vikash Kumar.
\newblock Dexterous manipulation with deep reinforcement learning: Efficient, general, and low-cost.
\newblock In \emph{International Conference on Robotics and Automation, {ICRA} 2019, Montreal, QC, Canada, May 20-24, 2019}, pages 3651--3657. {IEEE}, 2019.
\newblock \doi{10.1109/ICRA.2019.8794102}.
\newblock URL \url{https://doi.org/10.1109/ICRA.2019.8794102}.

\bibitem[Zhu et~al.(2020)Zhu, Yu, Gupta, Shah, Hartikainen, Singh, Kumar, and Levine]{zhu20r3l}
Henry Zhu, Justin Yu, Abhishek Gupta, Dhruv Shah, Kristian Hartikainen, Avi Singh, Vikash Kumar, and Sergey Levine.
\newblock The ingredients of real world robotic reinforcement learning.
\newblock In \emph{8th International Conference on Learning Representations, {ICLR} 2020, Addis Ababa, Ethiopia, April 26-30, 2020}. OpenReview.net, 2020.
\newblock URL \url{https://openreview.net/forum?id=rJe2syrtvS}.

\end{thebibliography}
\end{document}